\documentclass[final,1p,times,twocolumn]{elsarticle}

\usepackage{amssymb}

\usepackage{graphicx}
\usepackage{caption} 
\usepackage{amsfonts,amssymb, amsthm, amsmath}
\usepackage{subfigure}
\usepackage{color}
\usepackage{rotating}
\usepackage{multirow}
\usepackage{multicol}
\usepackage[linesnumbered,ruled, vlined]{algorithm2e}
\usepackage{algpseudocode}
\usepackage{hyperref}

\newcommand{\Sref}[1]{Section~\ref{#1}}
\newcommand{\tref}[1]{Tab.~\ref{#1}}
\newcommand{\eref}[1]{Eq.~(\ref{#1})}
\newcommand{\fref}[1]{Fig.~\ref{#1}}

\newcommand{\Aref}[1]{Algorithm~\ref{#1}}

\newcommand{\qiankun}[1]{\textcolor{black}{#1}}

\usepackage{amsmath}
\DeclareMathOperator*{\argmax}{arg\,max}

\journal{Journal of Neurocomputing}
\begin{document}
	
	\begin{frontmatter}
		
		
		
		\title{Online Multi-Object Tracking with Unsupervised Re-Identification Learning and Occlusion Estimation}
		\author[aff1]{Qiankun Liu\fnref{equ1}}
		\fntext[equ1]{Equal contribution.}
		\ead{liuqk3@mail.ustc.edu.cn}
		\author[aff2]{Dongdong Chen\fnref{equ1}}
		\ead{cddlyf@gmail.com}
		\author[aff1]{Qi Chu\corref{cor1}}
		\cortext[cor1]{Corresponding author. }
		\ead{qchu@ustc.edu.cn}
		\author[aff2]{Lu Yuan}
		\ead{luyuan@microsoft.com}
		\author[aff1]{Bin Liu}
		\ead{flowice@ustc.edu.cn}
		\author[aff3]{Lei Zhang}
		\ead{leizhang@idea.edu.cn}
		\author[aff1]{Nenghai Yu}
		\ead{ynh@ustc.edu.cn}
		
		\affiliation[aff1]{organization={University of Science and Technology of China},
			city={Hefei},
			country={China}}
		\affiliation[aff2]{organization={Microsoft Cloud AI},
			city={Redmond},
			country={U.S.}}
		\affiliation[aff3]{organization={International Digital Economy Academy},
			city={Shenzhen},
			country={China}}

		\begin{abstract}
			Occlusion between different objects is a typical challenge in Multi-Object Tracking (MOT), which often leads to inferior tracking results due to the missing detected objects. The common practice in multi-object tracking is re-identifying the missed objects after their reappearance. Though tracking performance can be boosted by the re-identification, the annotation of identity is required to train the model. In addition, such practice of re-identification still can not track those highly occluded objects when they are missed by the detector. In this paper, we focus on online multi-object tracking and design two novel modules, the unsupervised re-identification learning module and the occlusion estimation module, to handle these problems. Specifically, the proposed unsupervised re-identification learning module does not require any (pseudo) identity information nor suffer from  the  scalability  issue.
			The proposed occlusion estimation module tries to predict the locations where occlusions happen, which are used to estimate the positions of missed objects by the detector. Our study shows that, when applied to state-of-the-art MOT methods, the proposed unsupervised re-identification learning is comparable to supervised re-identification learning, and the tracking performance is further improved by the proposed occlusion estimation module.
			
		\end{abstract}
		
		
	\end{frontmatter}
	
	
	
	\section{Introduction}	
	\label{sec:introduction}
	Multi-Object Tracking (MOT) is a fundamental computer vision task with a wide range of applications, including autonomous driving, robot navigation and video analysis. Benefiting from the advance of object detection \cite{felzenszwalb2009object, ren2016faster, law2018cornernet, zhou2019objects}, the tracking-by-detection paradigm has become popular for MOT in the past decade. Though great performance has been achieved recently \cite{zhang2020fairmot, zhou2020tracking, stadler2021improving,zheng2021improving,shuai2021siammot}, occlusion between objects still remains challenging for MOT.
	
	In MOT scenarios, an object may be missed by the detector due to heavy occlusion, and then reappear after a short while. In order to identify such reappeared objects, re-identification (Re-ID) is often used to associate these reappeared objects with existing tracklets. Most existing MOT works
	\cite{wojke2017simple, liugsm, bergmann2019tracking, liu2021multi}
	adopt an independent Re-ID model to learn discriminative representations for objects, which introduces extra high computational cost since each object needs to be cropped out and fed into the pre-trained Re-ID model. To achieve real-time tracking, some works try to share the Re-ID feature computation with the backbone in anchor-based detector \cite{liu2019real, wang2019towards} or point-based detector \cite{zhang2020fairmot} by introducing an extra Re-ID branch that is parallel to the detection branch. Thanks to the sharing of feature maps between different branches, such methods can enable tracking multiple objects in a real-time way. 

	However, these methods
	\cite{zhu2018online, wojke2017simple, wang2019towards, zhang2020fairmot} still suffer from the scalability issue in the Re-ID representation learning. For example, \cite{zhu2018online, wang2019towards, zhang2020fairmot} combine several existing tracking and human detection (or Re-ID) datasets together and then learn the Re-ID representation by classifying each identity appeared in the combined dataset as one class (pseudo identity label). Such classification methods may work well for small datasets,
	but will encounter the learning difficulty when the identity number is huge, because the dimension of the sibling classification layer (fully connected layer) is linearly proportional to the identity number. More importantly, such supervised Re-ID module learning requires the annotation of identities, which is highly expensive and unscalable. 
	
	\begin{figure*}[t]
		 \includegraphics[width=1.0\columnwidth]{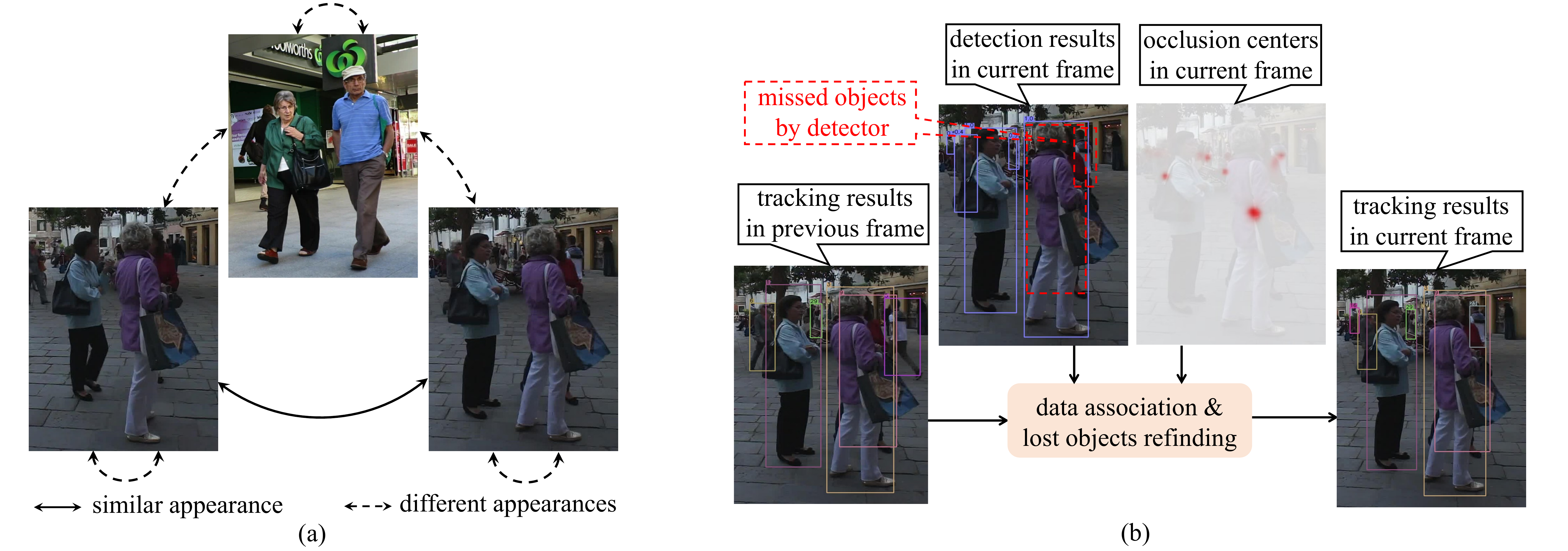} 
		\caption{(a) The left and right are two adjacent frames from a video while the middle is an image from another scene. An apparent observation is that objects with the same identity in adjacent frames share the similar appearance and objects in different scenes (or within the same image) have different identities and appearances. (b): The left and right are tracking results in previous and current frames, and the middle are the detected objects and occlusion centers in current frame. Some lost objects that are missed by detector can be tracked with the help of predicted occlusion centers. Please refer to \Sref{section_lost_object_refinding} for more details about lost objects refinding.
		}
		\label{figure_motivation}
	\end{figure*}

	To address this problem, we first propose a new Re-ID module learning mechanism. It adopts an unsupervised matching based loss between two frames (images)
	rather than the supervised classification loss used in \cite{zhu2018online, wang2019towards, zhang2020fairmot}. 
	This is based on the observation that objects with the same identity in adjacent frames share the similar appearance and objects in different scenes (or within the same image) have different identities and appearances,
	which is shown in \fref{figure_motivation} (a). Compared to the aforementioned methods, this newly proposed unsupervised Re-ID learning mechanism has two merits: 1) it does not need any (pseudo) identity annotation; 2) The matching based loss is irrelevant to the number of identities, thus can be directly trained on massive video-based data that with large number of identities.
	In addition, the image-based data can also be used for training if we treat two augmentations of one image as the adjacent frames.
	
	Though the Re-ID module can re-identify the reappeared objects after their short-term disappearance, how to proactively track the objects with highly occlusion is still challenging. This is because the severely occluded objects are easily missed by the detector, as shown in \fref{figure_motivation} (b). For example, in anchor-based detectors \cite{ren2016faster, he2017mask},  the Non-Maximum Suppression (NMS) module will remove highly overlapped boxes. In point-based detectors \cite{law2018cornernet, zhou2019objects}, as the object centers are invisible for occluded objects, it is also difficult to learn reliable center point-based features. Recently, how to address the missing detection issue caused by occlusion has attracted lots of attention. Some initial attempts \cite{zhu2018online, chi2020pedhunter, chu2020detection, zhu2020crowded} emerge, including detecting visible parts of an object \cite{zhu2018online, chi2020pedhunter}, using one proposal for multi-prediction \cite{chu2020detection}, and using paired anchors for one detection \cite{zhu2020crowded}. 	
	
	Different from existing methods, we propose a novel occlusion estimation module to predict whether two objects are occluded. Specifically, an occlusion map which shows all possible occlusion locations in the current frame is predicted. By further combing the status of existing tracklets, we finally design a lost object refinding mechanism to find the occluded objects back.
	
	To evaluate the effectiveness of the above two modules, we conduct extensive experiments by integrating them with different existing state-of-the-art MOT methods. For example, by replacing the supervised classification based Re-ID module in FairMOT \cite{zhang2020fairmot}, the unsupervised Re-ID learning module can still achieve comparable results on the MOT Challenge datasets\cite{milan2016mot16, dendorfer2020mot20} but neither needs any identity annotation nor suffers from any scalability issue. By integrating the occlusion estimation module, both FairMOT \cite{zhang2020fairmot} and CenterTrack \cite{zhou2020tracking} can handle the occlusion better and achieve the performance gain.
	
	To summarize, our contributions are three-fold as below: 
	\begin{itemize}
		\item We propose a novel unsupervised Re-ID learning module without using any identity information. It can be trained on video-/image-based data, and also has better scalability to datasets that with massive identities. 
		\item We propose a new occlusion estimation module, which can effectively recognize and track occluded objects when they are missed by the detector by estimating the occlusion location.  
		\item Both the unsupervised Re-ID learning and occlusion module can be applied to existing MOT methods in a natural way. Experimental results demonstrate the effectiveness of the proposed method.
	\end{itemize}
	
	The rest of this paper is organized as follows. In \Sref{section:related_work}, we review some related works in terms of MOT, person re-identification, and occlusion handling. Then in \Sref{section_method}, we elaborate the details of the two newly proposed modules, and the designed lost object refinding mechanism. To demonstrate the effectiveness, extensive experiment and ablation analysis are conducted in \Sref{section_experiments}. Finally, we conclude our work in \Sref{section_conclusion}. 
	
	\section{Related Work}
	\label{section:related_work}
	In this section, we first provide a brief overview about the popular tracking-by-detection paradigm for MOT, and then introduce the re-identification for data association in MOT as well as existing occlusion handling mechanisms in object detection and tracking.
	
	\subsection{Tracking-by-detection}
	Most existing MOT frameworks follow a tracking-by-detection paradigm thanks to the advances of object detectors \cite{ren2016faster, he2017mask, law2018cornernet, zhou2019objects}. Specifically, an object detector is used to detect objects in each frame, then a subsequent tracker is utilized to associate the objects across different frames. In terms of temporal information usage, existing MOT methods can be categorized into online
	\cite{wojke2017simple, bergmann2019tracking, liugsm, zhou2020tracking, zhang2020fairmot} and offline methods \cite{braso2020learning, hornakova2020lifted}. Online methods process video sequences frame-by-frame and track objects by only using information up to the current frame. By contrast, offline methods process video sequences in a batch and can even utilize the whole video information. From the network structure perspective, they can be further categorized into separate modeling 
	\cite{wojke2017simple,bergmann2019tracking,liugsm, braso2020learning, hornakova2020lifted} and joint modeling methods \cite{zhou2020tracking, zhang2020fairmot, wang2019towards}. 
	In separate modeling methods, the tracker is independently trained and assumes the detection results are available in advance. In joint modeling methods, the tracker is jointly trained with the detector by sharing the same feature extractor backbone. Therefore, they are often more efficient than the separate modeling methods. Both the newly proposed Re-ID module and occlusion estimation module can be naturally integrated into the online tracking-by-detection MOT system and jointly learned with the detector.
	
	\subsection{Re-identification for Association}
	Learning discriminative representations for objects is crucial to identity association in tracking. The representation can be used to re-identify lost objects that reappear after disappearing for a while. Early methods \cite{wojke2017simple, bergmann2019tracking, liugsm, babaee2019dual} crop the image patch of a detected object, resize and feed it into a separate Re-ID model. It is inevitably time-consuming since the feature representation of different objects has to be computed independently. To reduce the computation, some works attempt to share the Re-ID feature computation with the backbone in anchor-based detector \cite{liu2019real, wang2019towards, voigtlaender2019mots, porzi2020learning} or point-based detector \cite{zhang2020fairmot} by introducing an extra Re-ID branch that is parallel to detection branch. The common practice in MOT to train the Re-ID module is to classify each identity into one class \cite{zhu2018online, karthik2020simple, zhang2020fairmot, wang2019towards}. There are two fundamental weaknesses of such methods: 1) the Re-ID module is less scalable especially when the amount of identities is huge, because the classifier takes up a lot of memory. For example, FairMOT \cite{zhang2020fairmot} performs about 339K classification task to train the Re-ID module. 2) the training of Re-ID module needs to be supervised by identity information. For example, several datasets dedicated for Re-ID are adopted in \cite{zhang2020fairmot, wang2019towards}. However, the acquisition of well annotated data costs a lot.
	
	Despite the advance in supervised Re-ID learning \cite{fu2020improving,li2021triplet,sun2021visible,liu2021prgcn}, some works for unsupervised Re-ID learning have been proposed \cite{fu2021unsupervised,wang2020cycas, karthik2020simple, zhang2020fairmot, wu2020tracklet, fan2018unsupervised}.
		These works can be divided into two categories: pseudo identity based \cite{karthik2020simple, zhang2020fairmot, wu2020tracklet,fan2018unsupervised} and identity free methods \cite{wang2020cycas}. The proposed method is also an identity free method. For the former category, pseudo identities can be obtained by clustering \cite{wu2020tracklet, fan2018unsupervised} or tracking \cite{karthik2020simple}. However, the errors may accumulate and it is challenging to estimate the number of pseudo identities while clustering, and a trajectory of an object breaks into several short trajectories easily while tracking. For the latter category, the correspondence between adjacent frames is used \cite{wang2020cycas}. However, the birth and death of objects are not handled and the relation between objects within one frame is also not exploited.
	Inspired by these works, we propose to learn Re-ID representation in an unsupervised and matching based loss without using any (pseudo) identity information. It is built upon the observation that objects with the same identity in adjacent frames share the similar appearance and objects in different scenes (or within the same image) have different identities and appearances. Different from the works in \cite{wang2020cycas}, our method
	1) introduces a placeholder to handle the birth and death of objects and 2) makes usage of the information that objects in different scenes (or within the same image) have different identities.
	Besides, since our matching based loss is irrelevant to the number of identities, it does not suffer from the scalability issue and can be directly trained on massive data.

	\subsection{Occlusions between Objects}
	Handling highly occluded objects is a challenging task for both detection \cite{chu2020detection, zhu2020crowded, chi2020pedhunter} and tracking \cite{chu2017online,zhu2018online,chu2019online, chu2020dasot,xu2019spatial,liugsm}. Some anchor-based methods are proposed to handle occlusion \cite{chu2020detection, zhu2020crowded, chi2020pedhunter} in detection, including
	predicting multiple instances for one proposal~\cite{chu2020detection} or detecting one pedestrian with a pair of anchors~\cite{zhu2020crowded} (one is for head and another is for full body). However, both works require a carefully designed NMS algorithm for post processing. In \cite{chi2020pedhunter}, a mask-guided module is proposed to force the detector to pay attention to the more visible head part and thus detect the whole body of a pedestrian. 
	Different from these methods, we propose a new occlusion estimation mechanism upon key-point based detection, which detects the locations where occlusions happen and finds the missed objects caused by detector by combining the tracking status of existing tracklets.
	
	Instead of handling occlusions in the detection stage, there also exist some methods \cite{chu2017online,zhu2018online,chu2019online, chu2020dasot,xu2019spatial,liugsm} attempting to handle occlusions in the tracking stage for MOT task. The works in \cite{chu2017online,chu2020dasot,zhu2018online,chu2019online} utilize the single object tracking (SOT) method for MOT. In details, a SOT tracker is created and maintained for each object. Once an object is heavily occluded and missed by the detector, the position of it could be estimated by the corresponding SOT tracker. Such practice indeed is an extra detection stage with dedicated detectors (i.e., SOT trackers). The topology between different objects is also exploited to handle occlusions for MOT \cite{xu2019spatial,liugsm}. The hypothesis is that the topology between different objects in adjacent frames is invariant, which is positive to the association of objects, especially when some objects are partially occluded. In addition, the position of a lost object is estimated using the positions of its tracked neighbors \cite{liugsm} based on the topology among them in previous frame. 

	Different from these works, our method detects the locations of occlusion in a frame, and utilizes them to refind the missed objects while tracking online. More detail, if an object is missed by the detector, then it is likely to be heavily occluded by other objects. The detected occlusion locations could be used as the prior information to find it back.

	Occlusion is also one critical issue in SOT task \cite{tu2021novel, dong2020clnet}. Trying to find the visible region of the single object is the main focus in SOT. However, our method detects the overlapped region between different objects.
	
\section{Method}
	\label{section_method}
	As mentioned above, our paper proposes two key modules for existing multiple object tracking systems. One is the unsupervised Re-ID module learning mechanism, which is competitive to existing supervised counterparts and has better scalability. Another is the occlusion estimation module, 
	which predicts occlusion map to find the occluded objects back.
	In this section, we first elaborate the details of these two modules, and then show how to naturally integrate them with existing MOT systems.

	\subsection{Unsupervised Re-ID Learning}
	The proposed unsupervised Re-ID learning can be trained on both video-based and image-based data. \qiankun{Note that it is only used in the training stage. Once the training procedure is finished, the trained models can be directly used to extract discriminative features for different objects, which is the same as existing supervised counterparts.}
	For better understanding, we start with the learning from video-based data, then illustrate learning from image-based data. 
	
	\subsubsection{Learning From Video-Based Data}
	Let $I^t \in \mathbb{R}^{W \times H \times 3}$ be the $t$-th frame from one video and  $\boldsymbol b_i^t = (x^t_{i_l}, y^t_{i_t},$ $x^t_{i_r}, y^t_{i_b})$ be the groundtruth bounding box of object $i$ in frame $I^t$. $W,H$ are the width and height of the frame, and $(x^t_{i_l},y^t_{i_t}), (x^t_{i_r}, y^t_{i_b})$ are the coordinates of the top-left and bottom-right corners respectively. For Re-ID representation learning, we denote the appearance feature for object $i$ in frame $I^t$ as $f_i^t \in \mathbb{R}^D$, where $D$ is the dimension of the appearance feature vector. Our unsupervised Re-ID representation learning mechanism is general in how $f_i^t$ is calculated as long as it is differentiable. Possible solutions include cropping the image patch based on the given bounding box and feeding the cropped image patch into an extra Re-ID network like \cite{wang2020cycas, bergmann2019tracking, liugsm}, extracting the ROI based appearance features by sharing the same backbone as the detector like \cite{wang2019towards, voigtlaender2019mots}, and extracting the center point based appearance features like \cite{zhang2020fairmot}. The usage of annotated bounding boxes is equivalent to traditional Re-ID task in which the well cropped image patches are provided. Even though, no identity information is used in the proposed method.  In the following, the superscript $t$ of appearance feature is omitted for simplicity.

	\begin{figure*}[t]
		\centering
		\includegraphics[width=1.0\columnwidth]{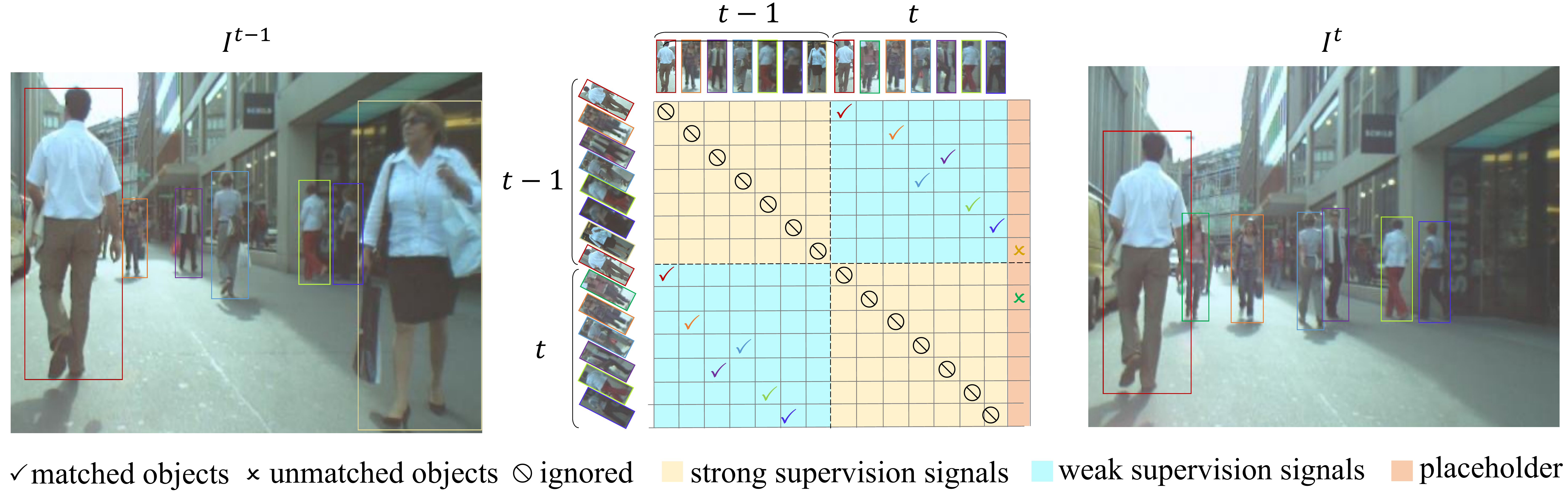} 
		\caption{The proposed un-supervised Re-ID learning method. Left and right are the two adjacent frames and the objects. Middle is the desired assignment results. For a better viewing, the identities of objects are encoded by color. However, the identity information is unused in our method. Two types of supervision signals are exploited. 1) Strong supervision signals: objects within the same frame should not be matched with each other. 2) Weak supervision signals: objects in one frame are likely to be matched with objects in another frame.}
		\label{fig_reid_learning}
	\end{figure*}

	Given two adjacent frames $I^{t-1}$ and $I^{t}$, let $i \in \{0,...,N^{t-1}-1, ..., N^{t-1}+N^{t}-1\}$ be the index of all objects in both two frames, where $N^t$ is the number of objects in frame $I^t$. The first $N^{t-1}$ objects are from frame $I^{t-1}$ and the rest are from frame $I^{t}$. We have two observations: 1) objects in the same frame have different identities; 2) an object is likely to appear in both adjacent frames. Accordingly, as shown in \fref{fig_reid_learning}, if we want to assign an object to another object, two types of supervision signals can be exploited: 1) objects within the same frame should not be matched with each other, which is regarded as strong supervision signal; 2) objects in one frame are likely to be matched with objects in another adjacent frame, which is weak supervision signal. In order to learn the Re-ID representation with such supervision signals, we first define a similarity matrix $S \in \mathbb{R}^{(N^{t-1}+ N^{t})\times(N^{t-1}+ N^{t})}$ that measures the similarity between each pair of objects, where:
	\begin{equation}
		S_{i,j}=
		\begin{cases}
			\frac{f_i \cdot f_j}{||f_i||_2  ||f_j||_2} & if \  i\neq j,\\
			- \infty & otherwise.
		\end{cases}
		\label{equ_cosine_distance}
	\end{equation}
	Obviously, $S_{i,j} = S_{j,i}$. The values in the diagonal of $S$ are set to negative infinity to avoid assigning an object to itself (\eref{equ_assign_without_placeholder}). In general, if objects $i$ and $j$ share the same identity, $S_{i,j} > 0$, otherwise, $S_{i,j} < 0$. The assignment matrix $M \in \mathbb{R}^{(N^{t-1}+ N^{t})\times(N^{t-1}+ N^{t})}$ can be obtained by applying row-wise softmax function to $S$ as:
	\begin{equation}
		M_{i,j} = \frac{e^{S_{i,j}T}}{\sum_{j} e^{S_{i,j}T}},
		\label{equ_assign_without_placeholder}
	\end{equation}
	where $T$ is the temperature of the softmax function.
	\qiankun{
	Consider the fact that the number of objects in adjacent frames (i.e., the size of $S$) could be various,  we follow the works in CycAs \cite{wang2020cycas} and set $T=2 {\rm log}(C+1)$, where $C=N^{t-1}+N^{t}$ is the number of columns in $S$. With this adaptive temperature, the maximum values in each row are almost equally highlighted/maximized even the size of $S$ varies.}	Since objects in the same frame have different identities, we can supervise the values in the top-left and right-bottom part of $M$ by a intra-frame loss:
	\begin{equation}
		L_{id}^{intra} = \sum_{0 \leq i,j \textless N^{t-1}}M_{i,j} + \sum_{N^{t-1} \leq i,j \textless N^{t-1}+N^{t}}M_{i,j}.
		\label{equ_id_loss_intra}
	\end{equation}
	This corresponds to the aforementioned strong supervision signal.
	To leverage the weak supervision signal, we first consider the ideal case where all the objects appear in both frames for better understanding. In this case, all the objects in the frame $I^{t-1}$ should be matched to the objects in the frame $I^{t}$ in a one-to-one manner. Then for each row in $M$, we encourage each object to be matched to another object with a high confidence by using the below inter-frame margin loss:
	\begin{equation}
		\begin{aligned}
			L_{id}^{inter} &= \sum_{i} \text{max}\{\max_{j',j' \neq j^*}M_{i,j'} + m - M_{i,j^*}, 0\},\\ &\text{where}\quad j^* = \argmax_j M_{i,j}.
			\label{equ_id_loss_inter}
		\end{aligned}
	\end{equation}
	This shares a similar spirit as the popular triple loss, i.e., the maximum matching probability $M_{i,j^*}$ is larger than the sub-maximum value by a pre-defined margin $m$ ($0.5$ by default).
	
	Besides the above margin loss, we further add another cycle constraint loss $L_{id}^{cycle}$ for $M$, which means the forward and backward assignment should be consistent with each other. In details, if an object $i$ in frame $I^{t-1}$ is matched with object $j$ in frame $I^{t}$, then the object $j$ in frame $I^{t}$ must be matched with object $i$ in frame $I^{t-1}$:
	\begin{equation}
		L_{id}^{cycle} =  \sum_{N^{t-1} \leq i \textless N^{t-1}+N^{t}, 0 \leq j \textless N^{t-1}} |M_{i,j}-M_{j,i}|.
		\label{equ_id_loss_consistent}
	\end{equation}
	
	Since two adjacent frames in video-based data often share some objects with the same identities, we call such two adjacent frames a positive sample for the Re-ID module training. The total loss for unsupervised Re-ID learning on such positive samples is:
	\begin{equation}
		L_{id}^{pos} = \frac{1}{N^{t-1}+N^{t}} (L_{id}^{intra} + L_{id}^{inter} + L_{id}^{cycle}).
		\label{equ_id_loss_pos}
	\end{equation}

	Unlike the above ideal case, an object in frame $I^{t-1}$ may disappear in frame $I^{t}$ (death of objects) and an object may appear in frame $I^{t}$ for the first time but invisible in frame $I^{t-1}$ (birth of objects) in a general case. 
	\qiankun{However, for each row in assignment matrix $M$, the inter-frame margin loss $L_{id}^{inter}$ will force the maximum value to be larger than the other values by a margin $m$, which is unsuitable when the corresponding object is disappeared or newly appeared since it does not share the same identity with any one of the other objects.}
	To handle this issue, a new similarity matrix $S' \in \mathbb{R}^{(N^{t-1}+ N^{t})\times(N^{t-1}+ N^{t}+1)}$ is obtained by padding a placeholder column to $S$. All values in the padded placeholder column are the same, which is denoted as $p$. The detailed discussion on $p$ is presented in Experiments \Sref{section_experiments_unsupervised_reid_learning}.
	\qiankun{With the existence of placeholder column, the similarity scores between disappeared/newly appeared objects and other objects are encouraged to be learned smaller than $p$.}
	Let $M'\in \mathbb{R}^{(N^{t-1}+ N^{t})\times(N^{t-1}+ N^{t}+1)}$ be the assignment matrix by applying row-wise softmax function to $S'$ \footnote{\qiankun{In this case, $C=N^{t-1}+N^{t}+1$ for the calculation of temperature $T$.}}. Then we replace $M$ with $M'$ in \eref{equ_id_loss_intra},  \eref{equ_id_loss_inter} and \eref{equ_id_loss_consistent} for this general case. 
	In our implementation, the loss for this general case is adopted.

	\subsubsection{Learning From Image-Based Data}
	To train the proposed Re-ID module on image-based data, a straightforward way is to get two augmentations of one image and treat these two augmentations as adjacent frames like the video-based data. However, we find only using the above positive sample based loss does not perform very well, since objects in the two augmentations have very similar appearance, thus not strong enough in learning discriminative Re-ID features. Considering the fact that objects in two different static images usually have different identities, we further introduce a negative sample based loss $L_{id}^{neg}$ by treating two different static images from different scenes as a negative sample pair:
	\begin{equation}
		L_{id}^{neg} = \sum_{0 \leq i,j \textless N^{t-1}+N^{t}}M'_{i,j}.
		\label{equ_id_loss_neg}
	\end{equation}
	Similarly, in this formulation, we introduce the extra placeholder $p$ and encourage the cosine distance between the objects in the negative pair to be less than $p$, which also means that all objects should be assigned to the placeholder. 
	\qiankun{Note that the design of $L_{id}^{neg}$ shares the same spirit with the intra-frame loss $L_{id}^{intra}$, while the inter-frame margin loss $L_{id}^{inter}$ and cycle constraint  loss $L_{id}^{cycle}$ are not used for negative sample pairs.} 
	
	Therefore, the overall unsupervised Re-ID learning loss for the image based data is:
	\begin{equation}
		L_{id} = \frac{N^{pos}}{N^{pos}+N^{neg}} L^{pos}_{id} + \frac{N^{neg}}{N^{pos}+N^{neg}} L^{neg}_{id},
		\label{equ_id_loss}
	\end{equation}
	where $N^{pos}$ and $N^{neg}$ are the number of positive and negative samples in a batch. In our default setting, $\frac{N^{neg}}{N^{pos}}$ is set to 0.25.

	Although the Re-ID module can help re-identify reappeared objects after their short-term disappearance, it is inherently unable to track the occluded objects if they are not detected by the detector. To mitigate the issue caused by the missed detection, we propose an occlusion estimation module to predict whether any occlusion occurs and find lost objects back by combining the predicted occlusions and the tracking status of existing tracklets.

	\subsection{Occlusion Estimation Module}
	\label{section_occlusion_estimation_module}
	
	\begin{figure}[t]
		\centering
		\includegraphics[width=0.6\columnwidth]{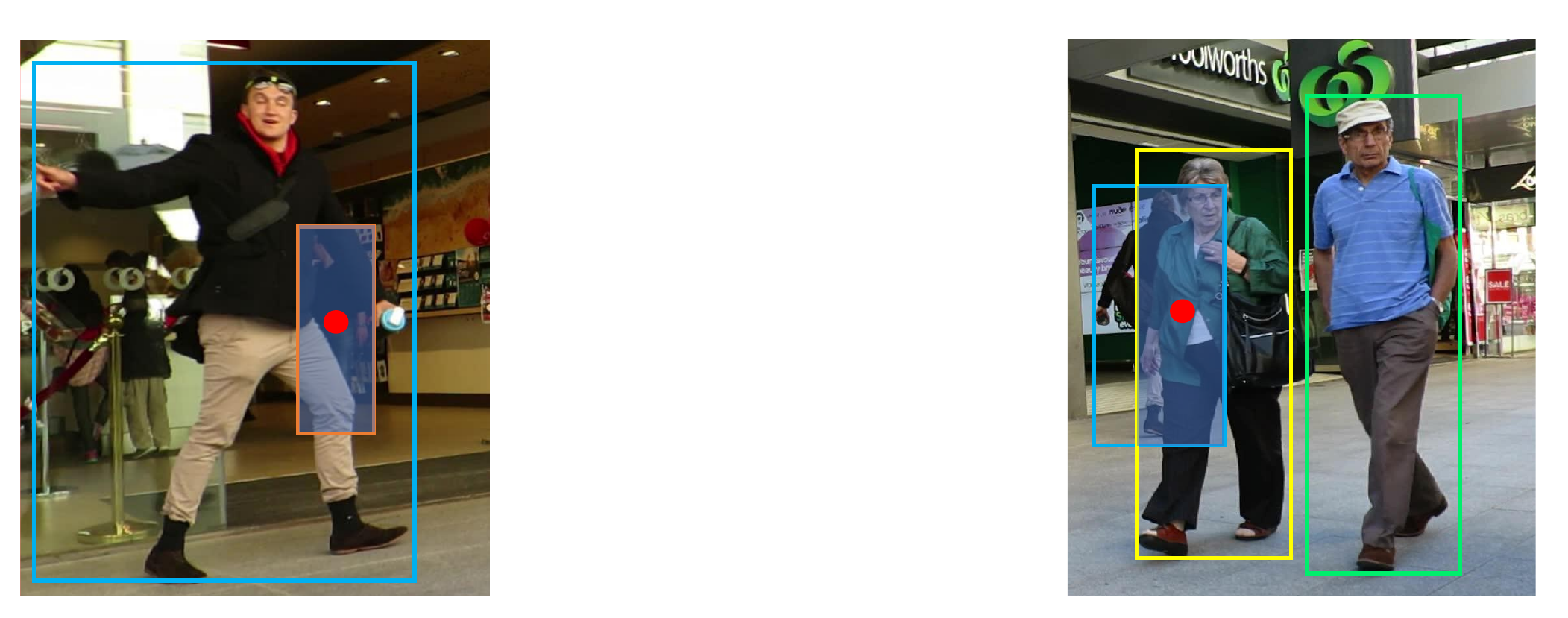}
		\caption{Typical occlusion cases. Translucent blue areas denote the positions where occlusions happen and red circles are the occlusion centers. Left: A small box covered by a larger box. Right: Two boxes overlapped with each other.}
		\label{figure_occlusion_defination}
	\end{figure}

	\subsubsection{Occlusion Detection}
	\label{section_occlusion_detection}
	Inspired by the work of key-point estimation \cite{law2018cornernet, zhou2019objects}, the locations of occlusion are treated as key-points and detected by key-point estimation. 
	Different from the above Re-ID module, the learning of the occlusion estimation module is designed in a supervised way. We automatically generate occlusion annotation based on the bounding boxes of objects, which are available in existing tracking datasets like MOT16 and MOT17 \cite{milan2016mot16}.
	
	First, we need to define when an occlusion occurs. Given the bounding box coordinates of two objects $i$ and $j$ within one frame, their overlapped region is defined as $\boldsymbol{o}_{ij} = \mathcal{O}(\boldsymbol{b}_i, \boldsymbol{b}_j) = (x_{ij_l}, y_{ij_t}, x_{ij_r}, y_{ij_b})$.
	Considering two typical occlusion examples as shown in \fref{figure_occlusion_defination}, 
	\qiankun{we define an indicator function $\mathcal{H}(\cdot)$ that indicates whether an occlusion is valid or not. Only when the overlapped region occupies a large portion of object $i$ or $j$, the occlusion $\mathbf{o}_{ij}$ is valid. Specifically:}
	\begin{equation}
		\mathcal{H}(\boldsymbol{o}_{ij})=\left\{
		\begin{array}{rc}
			1 & if \  \frac{\mathcal{A}(\boldsymbol{o}_{ij})}{{\rm min}(\mathcal{A}(\boldsymbol{b}_i), \mathcal{A}(\boldsymbol{b}_j))} > \tau,  \\
			0 & else ,
		\end{array}
		\right.
		\label{effectiveness of occlusion}
	\end{equation}
	where $\mathcal{A}(\cdot)$ is the function computing the area of a box, and $\tau$ is a hyper-parameter which is set as 0.7 by default. 
	In order to refind an occluded object back (Section \ref{section_lost_object_refinding}), we define the center point of $\boldsymbol o_{ij}$ as the occlusion location of two overlapped objects. The groundtruth occlusion map $Y$ is rendered by a 2D Gaussian kernel function based on  all the valid occlusions defined in \eref{effectiveness of occlusion} as:
	\begin{equation}
		Y_{x,y} = {\rm max}_{ij} \mathcal{G}(\boldsymbol{o}_{ij}, (x,y)), {\rm \ subject \  to \ } \mathcal{H}(\boldsymbol{o}_{ij})=1,
		\label{function_render_occ_center}
	\end{equation}
	where $\mathcal{G}(\boldsymbol{o}_{ij}, (x,y)) = {\rm exp} (- \frac{((x,y) - \lfloor \frac{\boldsymbol{p}_{ij}}{R} \rfloor)^2}{2\sigma_{\boldsymbol{o}_{ij}}^2})$ is the Gaussian kernel function, and $\boldsymbol{p}_{ij}=(\frac{x_{ij_l}+x_{ij_r}}{2}, \frac{y_{ij_t}+y_{ij_b}}{2})$ is the center point of occlusion $\boldsymbol{o}_{ij}$. 
	The standard deviation $\sigma_{\boldsymbol{o}_{ij}}$ of the Gaussian kernel is set to be relative to the size of $\boldsymbol{o}_{ij}$ following the definition in \cite{law2018cornernet}. In our implementation, we introduce an extra CNN head to obtain the predicted occlusion center heatmap $\hat Y \in \mathbb{R}^{\frac{W}{R} \times \frac{H}{R}}$. It is parallel to the detection head and shares the same backbone network. $R$ is the downsampling factor of the backbone network. Intuitively, the value $\hat Y_{x,y} \in [0,1]$ denotes the probability of an occlusion center that locates in $(x, y)$ and is supervised by:
	\begin{equation}
		L_{occ}^{cen} = 
		\sum_{x,y} \mathcal{L}(Y_{x,y}, \hat Y_{x,y}),
		\label{eq_occ_center_loss}
	\end{equation}
	where $\mathcal{L}(\cdot, \cdot)$ is a variant of focal loss function used in \cite{law2018cornernet} with two hyper-parameters $\alpha,\beta$ (default values are 2 and 4 respectively):
	\begin{equation}
		\mathcal{L}(y, \hat y) = \left\{
		\begin{aligned}
			&\quad \ -(1-\hat y)^\alpha {\rm log}(\hat y) \quad if \  y =1,  \\
			&-(1-y)^\beta (\hat{y})^{\alpha}{\rm log}(1-\hat{y}) \ \   else .
		\end{aligned}
		\right.
		\label{eq_focal_loss}
	\end{equation}

	Considering that $R$ is often larger than $1$, we take the inspiration from \cite{zhou2019objects} and add another CNN head to produce an offset heatmap $\hat \Lambda \in \mathbb{R}^{\frac{W}{R} \times \frac{H}{R} \times 2}$, which  can help compensate the quantization error in generating the occlusion center heatmap $Y$. The simple L1 loss is used to regress the center offset:
	\begin{equation}
		L_{occ}^{off}=
		\sum_{ij}|\hat{\Lambda}_{\lfloor \frac{\boldsymbol{p}_{ij}}{R} \rfloor} - (\frac{\boldsymbol{p}_{ij}}{R} - \lfloor \frac{\boldsymbol{p}_{ij}}{R} \rfloor)|.
		\label{eq_occ_off_loss}
	\end{equation}
	Need to note that the offset supervision is only given at the center locations. The overall occlusion estimation loss is:
	\begin{equation}
		L_{occ}=\frac{1}{\sum_{i,j} \mathcal{H}(\boldsymbol{o}_{ij})} (L_{occ}^{cen} + L_{occ}^{off}).
		\label{eq_occ_loss}
	\end{equation}

	\subsubsection{Lost Object Refinding}
	\label{section_lost_object_refinding}
	\qiankun{While tracking online, the occlusion estimation module is used to detect the possible occlusion locations, i.e., the center points of overlapped regions between different objects in a frame. 
	For severely occluded objects, they are easily missed by the detector (thus lost by the tracker). In such case, the corresponding occlusion locations can be used as the prior information to refind them. Specifically, given the set of existing tracklets in frame $t-1$ and the set of newly detected objects in frame $t$,  we match the newly detected objects with existing tracklets.
	If some tracklets cannot match with any newly detected objects, we treat them as potential lost tracklets/objects and try to find them back. The detailed tracking logic is elaborated in \Aref{algorithm_tracking_online}. Through the refinding of lost objects, the number of false negative objects could be reduced, leading to a higher tracking performance.}
	
	\begin{figure}[t]
		\centering
		\includegraphics[width=1.0\columnwidth]{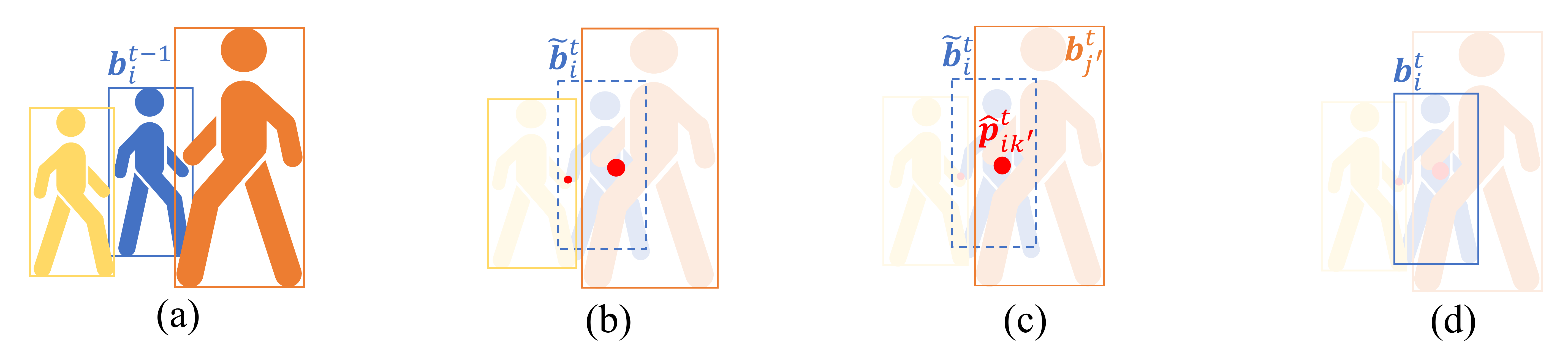}
		\caption{
			Illustration of lost object refinding. (a): Tracking results in the previous frame. (b): Predicted box $\tilde{\boldsymbol{b}}_{i}^{t}$ from $\boldsymbol{b}_{i}^{t-1}$ via motion, the overlapped boxes with $\tilde{\boldsymbol{b}}_{i}^{t}$, and the predicted occlusion center points that locate within $\tilde{\boldsymbol{b}}_{i}^{t}$. (c): The chosen occlusion center $\hat{\boldsymbol{p}}^{t}_{ik^{'}}$ and box $\tilde{\boldsymbol{b}}_{j^{'}}^{t}$ used for recovering $\boldsymbol{b}_{i}^{t}$. (d): Recovered box $\boldsymbol{b}_{i}^{t}$ for the lost object $i$.
		}
		\label{figure_lost_object_refinding}
	\end{figure}
	
	\qiankun{Once there exist some potential lost objects, we propose to find the lost objects back by using the predicted occlusion locations and the motion information of the corresponding tracklets, which can be estimated by Kalman filter. }
	In details, suppose we want to refind the lost object $i$ in $I^t$, its bounding box in $I^{t-1}$ is denoted as $\boldsymbol{b}_i^{t-1}=(x^{t-1}_{i_l}, y^{t-1}_{i_t},$ $x^{t-1}_{i_r}, y^{t-1}_{i_b})$. We first predict its location at $I^t$ via Kalman filter and denote the location as $\tilde{\boldsymbol{b}}_i^t=(\tilde x^t_{i_l}, \tilde y^t_{i_t}, \tilde x^t_{i_r}, \tilde y^t_{i_b})$. Then we search all the detected objects that possibly occlude $i$  by considering the estimated occlusion centers close to $\tilde{\boldsymbol{b}}_i^t$. The detailed search process is illustrated in \fref{figure_lost_object_refinding}. Specifically, for each box $\boldsymbol{b}_j^t$ that possibly overlapped with $\tilde{\boldsymbol{b}}_i^t$, 
	we first calculate the overlapped region as $\tilde{\boldsymbol{o}}_{ij} = \mathcal{O}(\tilde{\boldsymbol{b}}_i^t,\boldsymbol{b}_j^t)$. Then we get a score between $\tilde{\boldsymbol{o}}_{ij}$ and one of the predicted occlusion centers $\hat{\boldsymbol{p}}^t_{ik}=(\hat{x}^{t}_{ik}, \hat{y}^{t}_{ik})$  that locates within $\tilde{\boldsymbol{b}}_i^t$ using the aforementioned Gaussian kernel function.
	Finally, we choose the best matched pair by $(j^{'}, k^{'}) = {\rm argmax}_{j,k} \mathcal{G}(\tilde{\boldsymbol{o}}_{ij}^t, \hat{\boldsymbol{p}}^{t}_{ik})$. If $\mathcal{G}(\tilde{\boldsymbol{o}}_{ij^{'}}^t, \hat{\boldsymbol{p}}^{t}_{ik^{'}}) > \tau_o$ ($\tau_o = 0.7$ by default), 
	then object $i$ is likely to be occluded by object $j^{'}$, leading to missing detection. Suppose that the box $\boldsymbol{b}^t_{j^{'}}$ and occlusion center $\hat{\boldsymbol{p}}^t_{ik^{'}}$ are all correctly estimated and the size of object $i$ in adjacent frames keeps unchanged, the estimated box $\boldsymbol{b}_{i}^t$ for object $i$ can be calculated as:

	\begin{equation}
		\left\{
		\begin{array}{lr}
			x_{i_l}^{t} = \mathcal{F}(\tilde x^{t}_{i_l}, \tilde x^{t}_{i_r}, x^{t}_{j^{'}_l}, x^{t}_{j^{'}_r}, \hat{x}^{t}_{ik^{'}}), & \\ 
			y_{i_t}^{t} = \mathcal{F}(\tilde y^{t}_{i_t}, \tilde y^{t}_{i_b}, y^{t}_{j^{'}_t}, y^{t}_{j^{'}_b}, \hat{y}^{t}_{ik^{'}}), & \\ 
			x_{i_r}^{t} = x_{i_l}^{t} + \tilde x^{t}_{i_r} - \tilde x^{t}_{i_l}, & \\ 
			y_{i_b}^{t} = y_{i_t}^{t} + \tilde y^{t}_{i_b} - \tilde y^{t}_{i_t}, &
		\end{array}
		\right.
		\label{equation_lost_objet_refinding}
	\end{equation}
	where $\mathcal{F}(a_1, a_2, b_1, b_2, z)=$
	\begin{equation}
		\left\{
		\begin{array}{lc}
			2z - b_1 - (a_2 - a_1) &if \ a_1 \leq b_1 \ and \ a_2 \leq b_2,  \\
			z - (a_2 - a_1)/2 	   &if \ a_1 > b_1 \ and \ a_2 \leq b_2,  \\
			2z - b_2               &if \ a_1 > b_1 \ and \ a_2 > b_2,  \\
			a_1                    &else .
		\end{array} 
		\right.
		\label{equation_get_lost_coordinate}
	\end{equation}

	\begin{figure}[t]
		\centering
		\includegraphics[width=0.8\columnwidth]{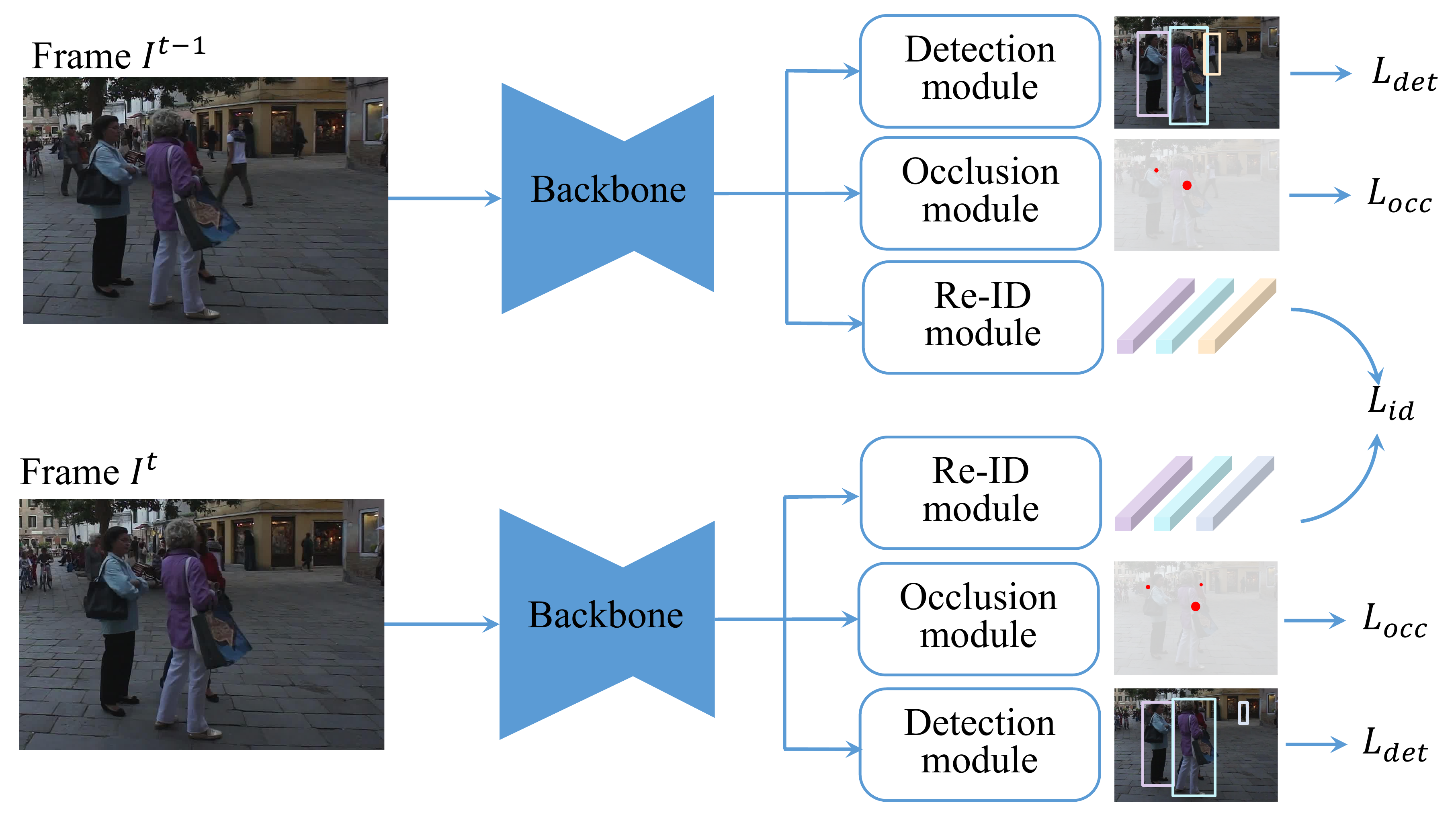}
		\caption{
			Illustration of applying of our unsupervised Re-ID module and occlusion module to
			FairMOT \cite{zhang2020fairmot}.
			While integrating,
			the occlusion module is added to be parallel with detection module which remains unchanged. Re-ID features from two frames are needed to train the Re-ID module.
		}
		\label{figure_applying_to_existing_methods}
	\end{figure}

	\begin{algorithm}[tp]
		\footnotesize
		\caption{Tracking logic between Two Consecutive Frames}
		\label{algorithm_tracking_online}
		\SetKwInOut{Input}{Input}
		\SetKwInOut{Output}{Output}
		\Input{$T^{t-1}=\{(\boldsymbol{b}^{t-1}_i, id_i, w_i, \boldsymbol{p}^{t-1}_i) \}_{i=1}^{N^{t-1}}$: cached trackltes in frame $t-1$ with box $\boldsymbol{b}_i^{t-1}$, identity $id_i$, number of consecutive frames $w_i$ being lost, center point $\boldsymbol{p}^{t-1}_i$. \newline
			$D^{t}=\{(\boldsymbol{b}^{t}_j, \boldsymbol{p}^{t}_j)\}_{j=1}^{N^{t}}$: detected objects in frame $t$ with box $\boldsymbol{b}_j^{t}$, center point $\boldsymbol{p}^{t}_j$. \newline
			$\hat P^{t} = \{\boldsymbol{\hat p}_{k}^{t}\}_{k=1}^{N_{occ}^{t}}$: predicted occlusion center points in frame $t$ 
		}
		\Output{$T^{t}$: cached tracklets in frame $t$. \newline
			$B^{t}$: tracking results in frame $t$.}
		\textbf{Step1}: Initialize as empty set \\
		{\quad\quad $T^{t} \leftarrow \emptyset$, $B^{t} \leftarrow \emptyset$ }\\
		\textbf{Step2}: Assign objects to tracklets, and get the assigned index pairs, lost tracklet and unassigned object indices \\
		\quad $\{(i_a,j_a)\}_{a=1}^{A}, \{i_l\}_{l=1}^{L}, \{j_u\}_{u=1}^{U} \leftarrow \Call{Assign}{T^{t-1}, D^{t}}$ \\
		\textbf{Step3}: Update tracklets with assigned objects\\
		\quad \For{$(i_a, j_a) \in \{(i_a,j_a)\}_{a=1}^{A}$}
		{
			$w_{j_a} \leftarrow 0$, $id_{j_a} \leftarrow id_{i_a}$ \\
			$T^{t} \leftarrow T^{t} \cup \{(\boldsymbol{b}^{t}_{j_a}, id_{j_a}, w_{j_a}, \boldsymbol{p}^{t}_{j_a} )\}$  \\
			$B^{t} \leftarrow B^{t} \cup \{\boldsymbol{b}^{t}_{j_a}, id_{j_a}\}$
		}
		\textbf{Step4}: Initialize new tracklets with unassigned objects \\
		\quad \For{$j_u \in  \{j_u\}_{u=1}^{U}$}
		{
			$w_{j_u} \leftarrow 0$, $id_{j_u} \leftarrow \Call{NewID}{\space}$ \\
			$T^{t} \leftarrow T^{t} \cup \{(\boldsymbol{b}^{t}_{j_u}, id_{j_u}, w_{j_u}, \boldsymbol{p}^{t}_{j_u})\}$ \\
			$B^{t} \leftarrow B^{t} \cup \{\boldsymbol{b}^{t}_{j_u}, id_{j_u}\}$
			\quad \\ 
		}
		\textbf{Step5:} Handle lost tracklets  \\ 
		\quad\For{$i_l \in  \{i_l\}_{l=1}^{L}$}
		{	
			\If(/*$\tau_{w}$ is the time window threshold*/){$w_{i_l} < \tau_{w}$}
			{
				
				/* select the occlusion point and box in current frame to recover the box for lost tracklet */ \\
				$j^{'}, k^{'} = {\rm argmax}_{j,k} \mathcal{G}(\mathcal{O}(\tilde{\boldsymbol{b}}_{i_l}^t, \boldsymbol{b}_{j}^{t}), \hat{\boldsymbol{p}}^{t}_{k})$ \\ 
				\uIf{$\mathcal{G}(\mathcal{O}(\tilde{\boldsymbol{b}}_{i_l}^t, \boldsymbol{b}_{j'}^{t}), \hat{\boldsymbol{p}}^{t}_{k'}) > \tau_o$}
				{   
					$\boldsymbol{b}^{t}_{i_l} \leftarrow$ get occluded object box  based on \eref{equation_lost_objet_refinding}\\
					$T^t \leftarrow T^t \cup \{(\boldsymbol{b}^{t}_{i_l}, id_{i_l}, w_{i_l}, \boldsymbol{p}^{t}_{i_l})\}$  \\
					$B^t \leftarrow B^t \cup \{\boldsymbol{b}^{t}_{i_l}, id_{i_l}\}$ \\
				}
				\Else{
					$w_{i_l} \leftarrow w_{i_l} + 1$\\
					$T^t \leftarrow T^t  \cup \{(\tilde{\boldsymbol{b}}^{t}_{i_l}, id_{i_l}, w_{i_l}, \boldsymbol{p}^{t}_{i_l})\}$  \\
					
				}
			}
		}
		Return $T^t$, $B^t$.
	\end{algorithm}
	\subsection{Integration into Existing Methods}
	\label{section_applying_to_existing_methods}
	The above two modules can be naturally integrated into existing state-of-the-art MOT systems, such as \cite{wang2019towards, zhang2020fairmot,zhou2020tracking,liugsm}. In details, as long as the original MOT system has or is able to add the differentiable Re-ID feature learning part, the proposed Re-ID learning mechanism can be applied into it and allows large scale unsupervised Re-ID learning. For the occlusion estimation module, it is compatible with MOT systems that are equipped with the modern CNN detector. We can simply implement it by adding the occlusion estimation module as a parallel head to the original detection head and sharing the same CNN backbone. 
	
	In \fref{figure_applying_to_existing_methods}, we take the popular tracking framework FairMOT \cite{zhang2020fairmot} as an example and show the integrated framework. The original FairMOT has one point based detection module, and one supervised Re-ID module that is learnt by classifying each identity as one independent class, which needs costly Re-ID annotation and suffers from the aforementioned dimension explosion problem for huge identity number. We integrate the two proposed modules by changing its Re-ID learning mechanism and adding the occlusion estimation module as described above. In the following experiments, besides FairMOT, we also try to integrate our modules into CenterTrack \cite{zhou2020tracking}. Since CenterTrack does not have the Re-ID feature learning module, we only incorporate the occlusion estimation module into it.
	
	\section{Experiments}
	\label{section_experiments}
	\subsection{Implementation Details}
	The state-of-the-art methods FairMOT \cite{zhang2020fairmot} and CenterTrack \cite{zhou2020tracking} are both implemented based on the key-point based detector CenterNet \cite{zhou2019objects}. We integrate the proposed modules into them to demonstrate the effectiveness. The occlusion loss $L_{occ}$ is added to the detection loss of FairMOT and CenterTrack with the weight of $0.5$. The estimation branch for occlusion centers and offsets in the occlusion estimation module both consists of one $3 \times 3$ convolutional layer whose output is a 256-channel feature map and one $1 \times 1$ convolutional layer that produces the task-specific heatmap. Between these two layers, a ReLU activation function is adopted. For the occlusion center branch, the output heatmap $\hat Y$ is activated by the sigmoid function, while for the occlusion offset heatmap $\hat \Lambda$, no activation function is adopted. When replacing the supervised Re-ID learning in FairMOT \cite{zhang2020fairmot} with our unsupervised Re-ID learning, we directly substitute the original Re-ID loss in FairMOT with the loss $L_{id}$ in \eref{equ_id_loss} while keeping other unchanged. The dimension $D$ of Re-ID feature is set to $256$.
	
	By default, the Adam optimizer \cite{kingma2014adam} with the initial learning rate $1e-4$ is used. The models in FairMOT and CenterTrack are trained for $30$ and $70$ epochs respectively. For the positive samples of unsupervised Re-ID learning from video-based data, the adjacent frames are randomly sampled from consecutive $20$ frames.
	
	\subsection{Datasets}
	The proposed method is evaluated on the standard MOTChallenge datasets, including MOT16, MOT17 \cite{milan2016mot16} and MOT20 \cite{dendorfer2020mot20}.
	There are 7 training and other 7 testing videos in MOT16. MOT17 contains the same videos as MOT16 but with different annotations. MOT20 contains 4 training videos and 4 testing videos. The videos in MOT20 are captured in crowd scenes, which are quite different from those in MOT16 and MOT17. External dataset CrowdHuman \cite{shao2018crowdhuman} is adopted for pre-training. Note that pre-training on external dataset is a common practice in previous works
	\cite{sadeghian2017tracking, zhang2020fairmot, wang2019towards, wang2021multiple, zheng2021improving}. Besides the bounding box annotation for detection, identity information is also provided in MOT16, MOT17 and MOT20. However, the identity information is not used in our training process.
	
	We adopt the standard metrics of MOTChallenge 
	for evaluation, including:
	Multi-Object Tracking Accuracy (MOTA) \cite{bernardin2008evaluating},
	Multi-object Tracking Precision (MOTP) \cite{bernardin2008evaluating},
	ID F1 Score (IDF1),
	Mostly Tracked objects (MT),
	Mostly Lost objects (ML),
	Number of False Positives (FP),
	Number of False Negatives (FN),
	Number of Identity Switches (IDS) \cite{li2009learning} and number of Fragments (Frag).
	Some other metrics, including R1 and mAP, are also introduced for the evaluation of different Re-ID methods.
	
	\subsection{Ablation Studies}
	Without losing generality, we do ablation study on the MOT17 dataset for simplicity, following the work in FairMOT \cite{zhang2020fairmot} and CenterTrack \cite{zhou2020tracking}. Since no validation data is provided in the MOTChallenge, the common practice is to split each video in the training set into two half videos, the first part is for training and the second part is for validation \cite{zhang2020fairmot, zhou2020tracking, saleh2021probabilistic, wang2021multiple, wu2021track}. No external dataset is used if not specified.

	\subsubsection{Unsupervised Re-ID Learning}
	\label{section_experiments_unsupervised_reid_learning}
	In this sub-section, we conduct the ablation study for the unsupervised Re-ID learning module by integrating it into the MOT system FairMOT \cite{zhang2020fairmot}.

	\begin{table}
		\caption{Tracking results on the MOT17 validation set with respect to different Re-ID learning methods.  FairMOT\textsuperscript{\ref{ourself_trained_fairmot}} and FairMOT$_{\rm w/o}$ mean the results with or without the original supervised Re-ID method used in FairMOT.
			$\uparrow$ means the larger the better and $\downarrow$  means the smaller the better.
			Best results are shown in \textbf{bold} and highlighted with underline.
		}
		\setlength{\tabcolsep}{0.1pt}
		\centering
		\scriptsize
		\setlength{\tabcolsep}{0.75mm}{
			\begin{tabular}{c|c|c|c|c|c|c|c}
				\hline
				Trackers  &MOTA$\uparrow$ &IDF1 $\uparrow$ &MT$\uparrow$ &ML$\downarrow$ &FP$\downarrow$ &FN $\downarrow$ &IDS$\downarrow$ \\
				\hline
				FairMOT$_{\rm w/o}$ &65.8\% &61.0\% &133 &62 &\underline{\bf2620} &14735 &1098 \\
				FairMOT\textsuperscript{\ref{ourself_trained_fairmot}} \cite{zhang2020fairmot} &67.5\% &70.2\% &134 &\underline{\bf55} &2814 &\underline{\bf14263} &\underline{\bf492} \\
				FairMOT+CysAs \cite{wang2020cycas} &67.0\% &70.8\% &134 &60 &2631 &14681 &503  \\
				FairMOT+UTrack &\underline{\bf67.6\%} &\underline{\bf71.8\%} &\underline{\bf137} &63 &2621 &14388 &503  \\
				\hline
			\end{tabular}
		}
		\label{table_different_method_for_reid}
	\end{table}	
	
	\textbf{Comparison with other Re-ID learning methods: }
	We first compare different learning methods for the Re-ID feature module, including the proposed method (UTrack), the   unsupervised method CycAs \cite{wang2020cycas} and the supervised method in FairMOT \cite{zhang2020fairmot}.  CycAs utilizes the cycle assignment consistence to learn the Re-ID module, which is the latest identity free method for Re-ID learning. FairMOT treats each identity as a class and the Re-ID module is trained in a supervised classification manner. The detailed comparison results are shown in \tref{table_different_method_for_reid}. In order to demonstrate the effectiveness of the Re-ID module, the tracking results without Re-ID are also presented in the first row and denoted as FairMOT$_{\rm w/o}$. Note that, except the training method of the Re-ID module, all the other parts remain the same.

	Comparing the first row with the remaining three rows, we can find that MOTA, IDF1 and IDS are all greatly improved with the Re-ID module. For example, our UTrack improves MOTA and IDF1 by $1.8\%$ and $10.8\%$ respectively when the tracker is equipped with the Re-ID module trained by the proposed unsupervised method.
	Compared to the supervised Re-ID used in FairMOT\footnote{This tracker is trained by ourselves using its official code since the model is not available, which achieves the same MOTA, higher IDF1 and lower IDS.\label{ourself_trained_fairmot}}, our UTrack can achieve almost the same MOTA even without using any Re-ID supervision. Though FairMOT possesses a slightly lower IDS, our method UTrack performs better in IDF1 by $1.6\%$, demonstrating the effectiveness of the proposed unsupervised Re-ID learning method. 
	More importantly, our method does not suffer from the dimension explosion issue and is more friendly to the real large-scale MOT systems.
	Comparing our method UTrack with CysAs \cite{wang2020cycas}, both of whom are unsupervised methods, we find that both trackers achieve the same IDS, but our method performs better in terms of MOTA and IDF1. We attribute this to the introduction of the placeholder in the similarity matrix and the strong supervision signal $L_{id}^{intra}$ within the same frame (\eref{equ_id_loss_intra}).

	\begin{figure*}[t]
		\centering
		\subfigure[FairMOT]{
			\begin{minipage}[t]{0.33\linewidth}
				\centering
				\includegraphics[width=0.9\linewidth]{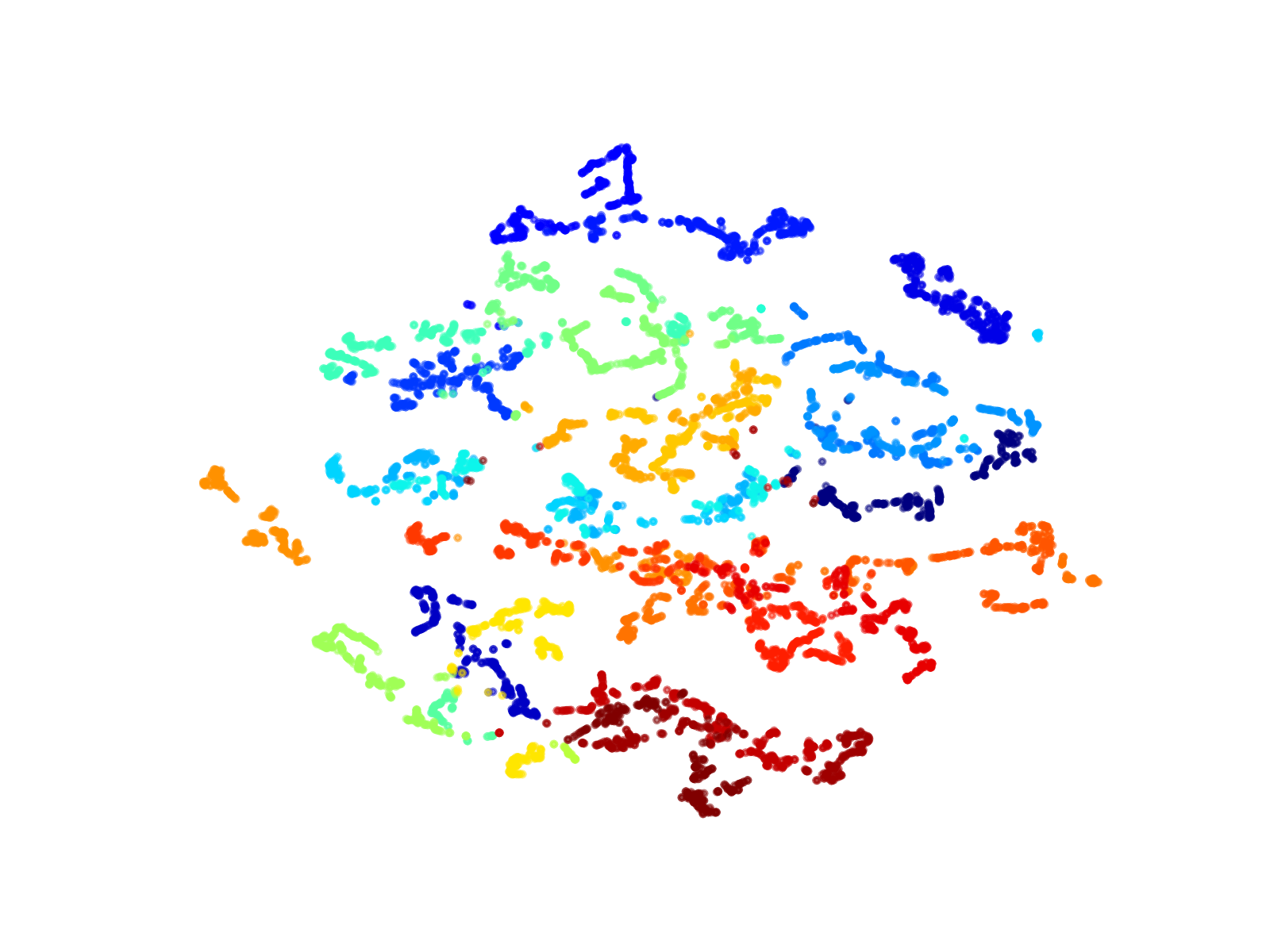}
			\end{minipage}%
		}%
		\subfigure[FairMOT+CycAs  \cite{wang2020cycas}]{
			\begin{minipage}[t]{0.33\linewidth}
				\centering
				\includegraphics[width=0.9\linewidth]{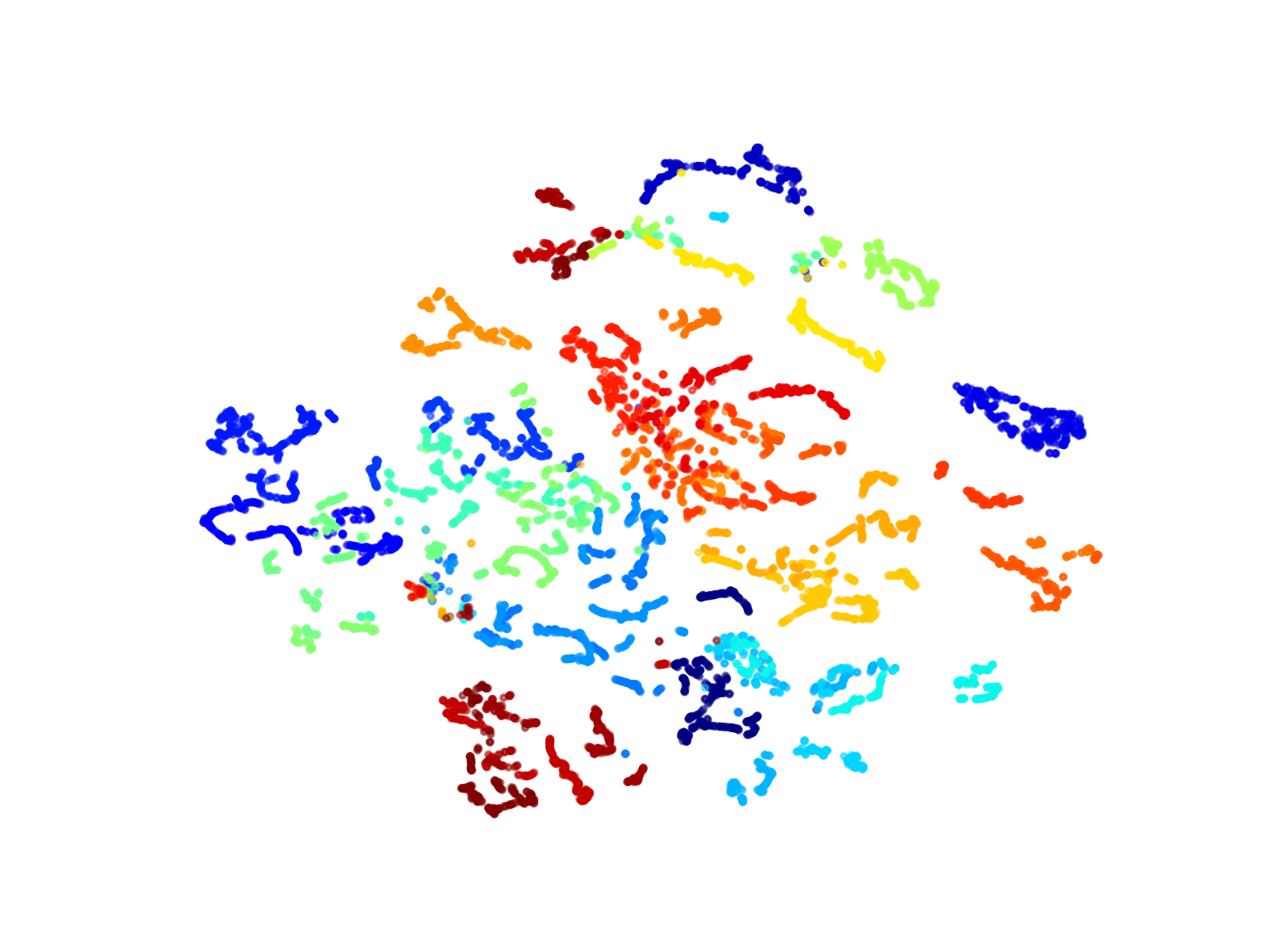}
			\end{minipage}%
		}%
		\subfigure[FairMOT+UTrack (Ours)]{
			\begin{minipage}[t]{0.33\linewidth}
				\centering
				\includegraphics[width=0.9\linewidth]{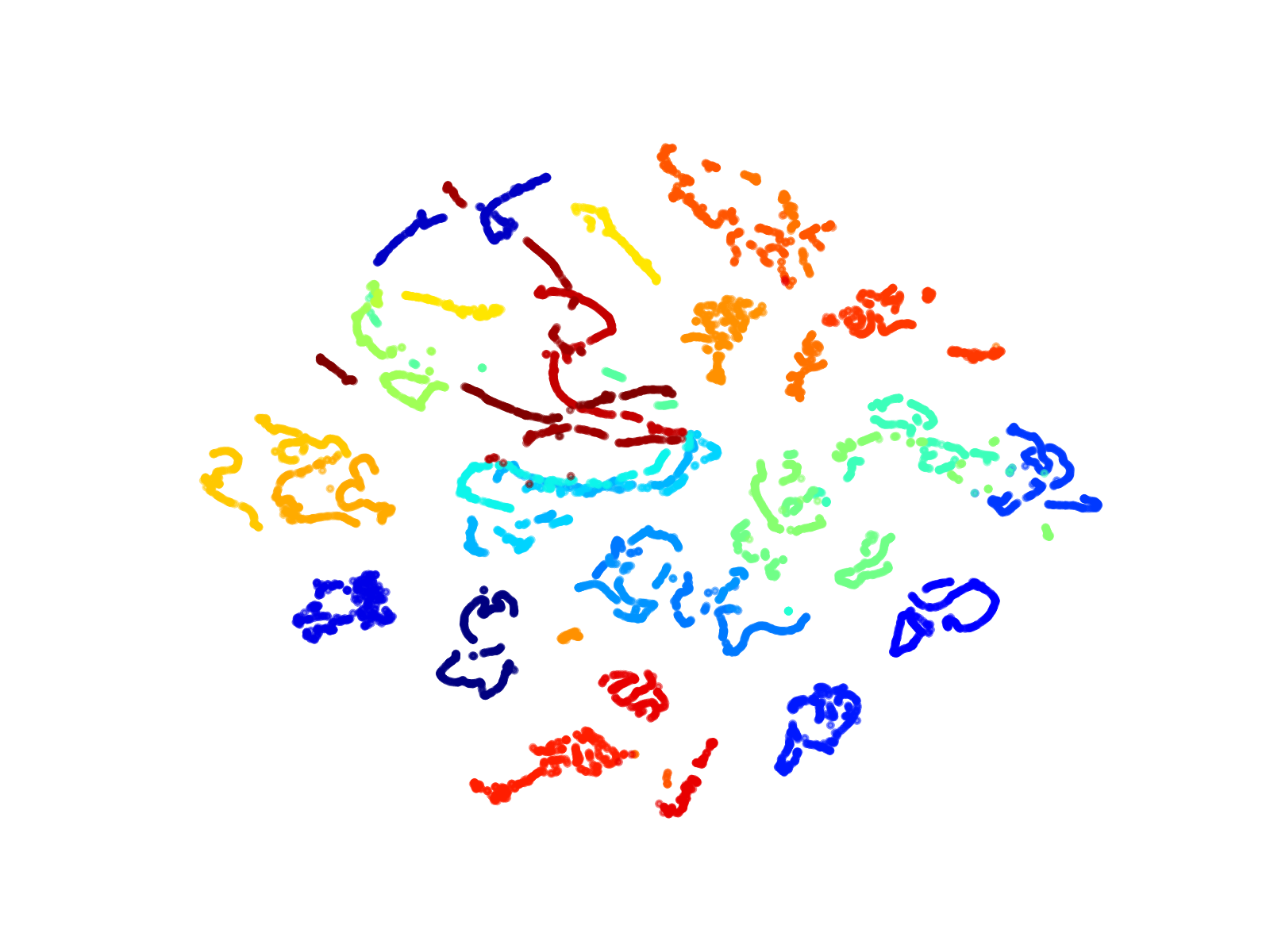}
			\end{minipage}
		}%
		\centering
		\caption{Visualized Re-ID features for identities in MOT17 validation set using t-SNE \cite{maaten2008visualizing}.
			From left to right are the features learnt by
			(a): the supervised method in FairMOT \cite{zhang2020fairmot}. (b): the unsupervised method CysAs  \cite{wang2020cycas}. (c): the proposed unsupervised method. Note that only the first 30 identities are presented here. Different colors indicate different identities.}
		\label{fig_id_feature_tsne}
	\end{figure*}

	\begin{figure*}
		\centering
		\includegraphics[width=1.0\columnwidth]{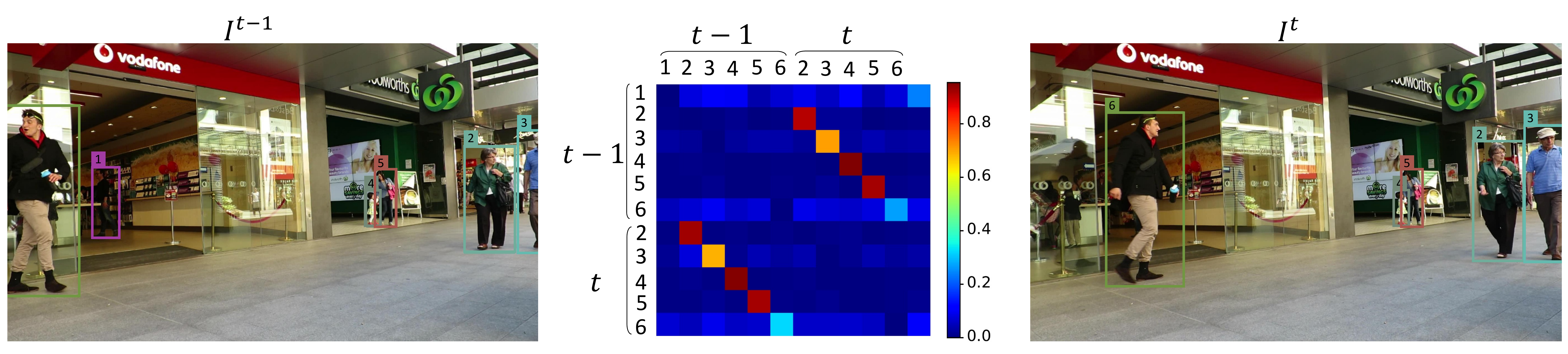}
		\caption{
			Illustration of assignment between two frames. Left and right are the detection results, in which the color of boxes and the numbers attached to the boxes indicate the identities. Middle is the assignment matrix $M'$.
		}
		\label{figure_id_feat_assignment}
	\end{figure*}

	In \fref{fig_id_feature_tsne}, we visualize the learned Re-ID features for different methods by t-SNE \cite{maaten2008visualizing}. As we can see, compared with the supervised method originally used in FairMOT \cite{zhang2020fairmot} and the unsupervised method CycAs \cite{wang2020cycas}, the features for the same identities are better grouped by the proposed unsupervised method. We further present the assignment between two frames in \fref{figure_id_feat_assignment}. As we can see, the object pair with the same identity achieves a much higher similarity score than the counterpart with different identities.

	\begin{table}
		\caption{Analysis of Re-ID loss on validation set. 
		}
		\setlength{\tabcolsep}{2pt}
		\centering
		\scriptsize
		\begin{tabular}{ccc|c|c|c|c|c|c|c}
			\hline
			$L_{id}^{inter}$ &$L_{id}^{intra}$ &$L_{id}^{cycle}$  &MOTA$\uparrow$ &IDF1 $\uparrow$ &MT$\uparrow$ &ML$\downarrow$ &FP$\downarrow$ &FN $\downarrow$ &IDS$\downarrow$ \\
			\hline
			\checkmark & & &67.3\% &69.9\% &134 &\underline{\bf58} &\underline{\bf2422} &14669 &592 \\
			\checkmark &\checkmark & &67.5\% &70.6\% &133 &63 &2492 &14566 &531 \\
			\checkmark &\checkmark &\checkmark &\underline{\bf67.6\%} &\underline{\bf71.8\%} &\underline{\bf137} &63 &2621 &\underline{\bf14388} &\underline{\bf503}  \\
			\hline
		\end{tabular}
		\label{table_analysis_of_reid_loss}
	\end{table}

	\textbf{Analysis of Re-ID loss:} We then do some ablation analysis on the Re-ID loss $L_{id}^{pos}$ in \eref{equ_id_loss_pos}. It consists of three components. 1) $L_{id}^{intra}$: the loss used to avoid assigning an object to another one that locates in the same frame. 2) $L_{id}^{inter}$: the loss that makes sure an object can be successfully matched with another object that locates in different frames or the placeholder. 3) $L_{id}^{cycle}$: the loss that constraints that the forward and backward assignment should be consistent with each other. Results are shown in \tref{table_analysis_of_reid_loss}. It can be seen that, the performance gain from different Re-ID losses in terms of MOTA is not that significant since MOTA is highly affected by the detection performance, i.e., FN and FP. But both IDS and IDF1 are improved by introducing more constraints on the Re-ID module, which demonstrates the effectiveness of each loss in \eref{equ_id_loss_pos}.

	\qiankun{\textbf{Discussion on the temperature $T$:} In our default settings, $T$ is adaptive to the number of objects. To show its superiority, we train several models with different fixed temperatures, the tracking results are shown in \tref{table_impact_of_temperature}. 
    As we can see, provided with larger temperatures ($T \ge 4$), the trackers achieve better IDF1 score but degraded MOTA score compared with those trackers equipped with smaller temperatures ($T\le 3$). With the help of adaptive temperature, the tracker obtains a better balance between IDF1 and MOTA scores.
	}	
	
	\begin{table}
		\caption{\qiankun{The impact of temperature $T$ on tracking performance. Evaluated on MOT17 validation set. $C$ is the number of columns in similarity matrix.}}
		\setlength{\tabcolsep}{0.1pt}
		\centering
		\scriptsize
		\setlength{\tabcolsep}{0.75mm}{
			\begin{tabular}{c|c|c|c|c|c|c|c|c}
				\hline
				Tracker & $T$  &MOTA$\uparrow$ &IDF1 $\uparrow$ &MT$\uparrow$ &ML$\downarrow$ &FP$\downarrow$ &FN $\downarrow$ &IDS$\downarrow$ \\ 
				\hline
				\multirow{6}*{FairMOT+UTrack} 
				& 1.0 &67.5\% &63.2\% &145 &\underline{\bf57} &2778 &13978 &788 \\
				& 2.0 &\underline{\bf67.9\%} &63.5\% &\underline{\bf147} &58 &2717 &13840 &789\\
				&3.0 &67.6\% &65.0\% &147 &\underline{\bf57} &2894 &\underline{\bf13791} &838 \\
				& 4.0 & 65.5\% &69.1\% &128 &62 &3186 &14874 &565 \\
				& 5.0 &65.4\% &70.1\% & 129 &59 &3130 &15059 &\underline{\bf502} \\
				& $2{\rm log}(C+1)$ &67.6\% &\underline{\bf71.8\%} &137 &63 &\underline{\bf2621} &14388 &503  \\
				\hline
			\end{tabular}
		}
		\label{table_impact_of_temperature}
	\end{table}

	\begin{table}
		\caption{Tracking results on the MOT17 validation set with respect to different settings for placeholder. 
		}
		\setlength{\tabcolsep}{4.4pt}
		\centering
		\scriptsize
		\begin{tabular}{c|c|c|c|c|c|c|c}
			\hline
			placeholder  &MOTA$\uparrow$ &IDF1 $\uparrow$ &MT$\uparrow$ &ML$\downarrow$ &FP$\downarrow$ &FN $\downarrow$ &IDS$\downarrow$ \\
			\hline
			w/o &67.4\% &71.1\% &\underline{\bf142} &62 &\underline{\bf2457} &14588 &557 \\
			zero  &66.9\% &70.7\% &138 &\underline{\bf60} &2685 &14735 &\underline{\bf494} \\
			mean  &\underline{\bf67.6\%} &\underline{\bf71.8\%} &137 &63 &2621 &\underline{\bf14388} &503  \\
			\hline
		\end{tabular}
		\label{table_discussion_on_placeholder}
	\end{table}

	\textbf{Discussion on the placeholder:}
	Finally, we have a discussion on the value of placeholder $p$. As mentioned before, the placeholder $p$ is introduced to handle the birth and death of objects, i.e., the newly appeared objects in $I^t$ and the objects appearing in $I^{t-1}$ but disappearing in $I^{t}$ should be assigned to the placeholder. Let $S_{i,j}$ be the cosine similarity between the Re-ID feature of objects $i$ and $j$. For sensible matching, $S_{i,j}$ should be greater than $p$ if object $i$ and $j$ have the same identity, otherwise $S_{i,j} < p$. 
	
	Taking into intuitive consideration that the cosine similarity between the Re-ID feature of two objects should be positive if they have the same identity, otherwise negative, it is straightforward to set $p = 0$. However, at the early training stage, we observe that the variance of the values in $S$ is small (about 0.015) and the cosine similarity between any pair of objects is around $0.75$. So it is hard for the model to handle the birth and death of objects well at the beginning if $p=0$. Therefore, we set $p$ as the dynamic mean of the values in $S$ except the diagonal values by default. Interestingly, 
	we observe that the mean of the values in $S$ is about $0$ 
	after convergence  when trained with this strategy.

	The results of three different placeholder settings are shown in \tref{table_discussion_on_placeholder}: without placeholder $p$, $p=0$, and $p$ as the dynamic mean. As we can see that the tracker achieves the best results in terms of MOTA, IDF1 and FN when the placeholder is set to the dynamic mean of similarity values. Compared to the setting without the placeholder, using placeholder can achieve much lower IDS, which demonstrates the effectiveness of the placeholder.

	\begin{table}
		\caption{Tracking results on the MOT17 validation set with the occlusion estimation module (OccE) integrated in CenterTrack \cite{zhou2020tracking} and FairMOT \cite{zhang2020fairmot} to show its effectiveness respectively. }
		\setlength{\tabcolsep}{0.4mm}
		\centering
		\scriptsize
		\begin{tabular}{c|c|c|c|c|c|c|c}
			\hline
			trackers  &MOTA$\uparrow$ &IDF1 $\uparrow$ &MT$\uparrow$ &ML$\downarrow$ &FP$\downarrow$ &FN $\downarrow$ &IDS$\downarrow$  \\
			\hline
			FairMOT+UTrack &67.6\% &71.8\% &137 &63 &\underline{\bf2621} &14388 &503 \\
			\textcolor{black}{FairMOT+UTrack+GSM \cite{liugsm}} &68.1\% &71.8\% &\underline{\bf164} & \underline{\bf49} &5095 &\underline{\bf11763} &\underline{\bf366} \\
			FairMOT+UTrack+OccE  &\underline{\bf68.5\%} &\underline{\bf72.0\%} &142 &57 &2840 &13797 &396 \\
			\hline
			CenterTrack &60.7\% &62.7\% &112 &76 &\underline{\bf2179} &18447 &564 \\
			\textcolor{black}{CenterTrack+GSM \cite{liugsm}} &61.5\% &63.9\% &\underline{\bf131} &\underline{\bf63} &4508 &\underline{\bf15943} &\underline{\bf254} \\
			CenterTrack+OccE  &\underline{\bf62.1\%} &\underline{\bf 64.6\%} &127 &68 &3372 &16583 &440 \\
			\hline
		\end{tabular}
		\label{table_occlusion_estimation_module}
	\end{table}	
	
	\subsubsection{Occlusion Estimation Module}
	\label{section_experiment_occlusion_estimation_module}
	To demonstrate the effectiveness of the proposed occlusion estimation module, we apply it to both FairMOT \cite{zhang2020fairmot} and CenterTrack \cite{zhou2020tracking}. 
	Another work of lost object refinding, GSM \cite{liugsm}, is also evaluated.
	As shown in \tref{table_occlusion_estimation_module}, with the help of the occlusion estimation module (OccE), many lost objects can be found back, leading to lower FN. Though the FP is slightly increased, the main tracking metric MOTA is still improved. In addition, more objects can be mostly tracked (MT), and fewer objects are mostly lost (ML). Besides, IDS are greatly reduced. 
	Compared with OccE, GSM introduces more FP, resulting a slightly lower MOTA. Though lower IDS is achieved by GSM, the construction and matching of graphs are time consuming.

	\begin{figure*}[]
		\centering
		\includegraphics[width=1.0\columnwidth]{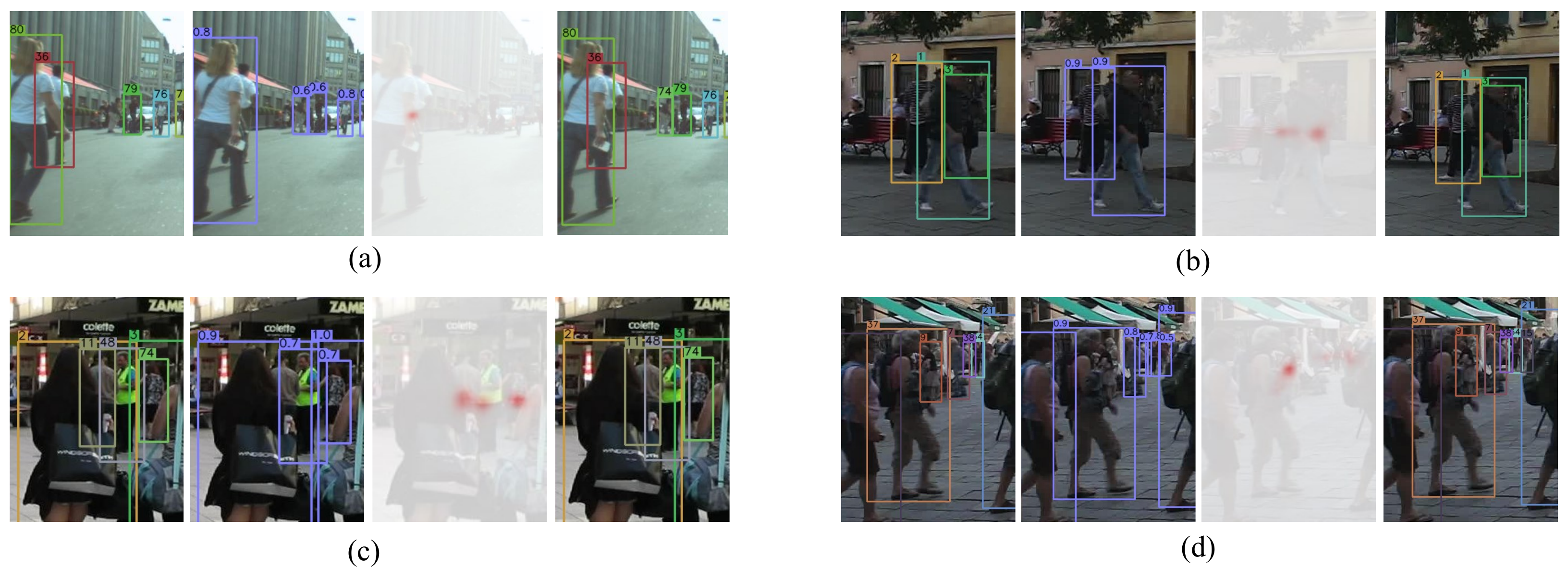}
		\caption{
			Some cases where lost objects are re-found by the proposed occlusion estimation module. For each case, from left to right are tracking results in the previous frame, detection results, predicted occlusions and tracking results in the current frame respectively. Specifically, objects 36, 3, 11 and 9 are found back for cases (a), (b), (c) and (d), respectively. Note that the images here are cropped from original images for better viewing. }
		\label{figure_refind_objects}
	\end{figure*}

	In \fref{figure_refind_objects}, some typical cases where occlusion happens are presented. For each case, the left to the right columns are the tracking results in the previous frame, the detection results and the predicted occlusions, as well as the tracking results in the current frame respectively. 
	As we can see, some occluded objects are missed by the detector and are challenging for existing MOT systems to track successfully. Without such refinding mechanism, they can only handle the detected objects and keep undetected boxes untracked. By integrating the proposed occlusion estimation module and the accompanying object refinding algorithm, we can refind the missed objects back and link them with existing tracklets.

	\begin{table}
		\caption{\qiankun{The impact of $\tau$ on tracking performance. Evaluated on MOT17 validation set.}}
		\setlength{\tabcolsep}{2pt}
		\centering
		\scriptsize
		\begin{tabular}{c|c|c|c|c|c|c|c|c}
			\hline
			trackers & $\tau$ &MOTA$\uparrow$ &IDF1 $\uparrow$ &MT$\uparrow$ &ML$\downarrow$ &FP$\downarrow$ &FN $\downarrow$ &IDS$\downarrow$  \\
			\hline
			\multirow{4}*{FairMOT+UTrack+OccE} 
			& 0.3 & 68.3\% &70.4\% &\underline{\bf145} &57 &3228 &\underline{\bf13501} & 412\\
			& 0.5 & 68.3\% &\underline{\bf72.0\%} &143 &\underline{\bf55} &3114 &13609 &426 \\
			& 0.7 &\underline{\bf68.5\%} &\underline{\bf72.0\%} &142 &57 &2840 &13797 &\underline{\bf396} \\
			& 0.9 & 68.4\% &71.8\% &141 &58 &\underline{\bf2758} & 13928 &397 \\
			\hline
		\end{tabular}
		\label{table_paramer_tau}
	\end{table}

	\qiankun{\textbf{Discussion on the threshold $\tau$:} The hyper-parameter $\tau$ in \eref{effectiveness of occlusion} controls the valid number of occlusions in a frame while training. 
	We evaluate the impact of $\tau$ on tracking performance by applying it to FairMOT \cite{zhang2020fairmot}. Results are shown in \tref{table_paramer_tau}. 
	Specifically, the tracker achieves more FN but less FP with a larger $\tau$. This is because fewer occlusions could be detected if the model is trained with a larger $\tau$, thus fewer lost objects could be found back. However, the overall tracking performance (MOTA) is not sensitive to the value of $\tau$ and we set it to 0.7 by default.}

	\subsubsection{Pre-Training on Image-Based Data}
	\label{section_pretraining_on_image_based_data}
	The proposed method can benefit from pre-training on image-based data. To demonstrate it, we use the CrowdHuman dataset \cite{shao2018crowdhuman} for pre-training. Need to note that, the original FairMOT \cite{zhang2020fairmot} also pre-trains their model on CrowdHuman and its Re-ID module is trained with pseudo identity labels, i.e., a unique identity is assigned to each annotated box, in a classification manner. There are about 339K boxes in CrowdHuman, so the pseudo identity number is massive, causing the number of parameters in the classifier to be even larger than the total number of parameters in the other modules (54.0M vs. 19.4M). By contrast, there are no extra parameters introduced in the proposed unsupervised Re-ID learning method.
	While training, two augmentations of an image are treated as a positive sample pair, and two different static images are sampled as a negative sample pair.
	
	\begin{table}
		\caption{Comparison of re-identification capability of different methods on the MOT17 train split by directly applying the pre-trained models on CrowdHuman \cite{shao2018crowdhuman} without fine-tuning.}
		\setlength{\tabcolsep}{10pt}
		\centering
		\scriptsize
		\begin{tabular}{c|c|c}
			\hline
			Trackers  &R1$\uparrow$ &mAP $\uparrow$ \\
			\hline
			FairMOT\cite{zhang2020fairmot} &42.9\% &25.4\% \\
			FairMOT+CysAs \cite{wang2020cycas} &54.8\% &32.9\% \\
			FairMOT+UTrack &\underline{\bf56.4\%} &\underline{\bf34.1\%} \\
			\hline
		\end{tabular}
		\label{table_pretrain_on_crowdhuman_no_ft_reid}
	\end{table}

	We first compare the re-identification capability of the proposed unsupervised Re-ID learning method UTrack, pseudo identity based method FairMOT \cite{zhang2020fairmot}, and the latest identity free method CysAs \cite{wang2020cycas} in \tref{table_pretrain_on_crowdhuman_no_ft_reid}. While evaluation, each tracklet in MOT17 train split is divided into two half parts. The first part is used as query and the rest part is used as gallery. As we can see, both CysAs and the proposed UTrack achieve much better results than FairMOT. Compared with CysAs, UTrack obtains 1.6\% higher R1 and 1.2\% higher mAP. We argue this to the introduction of placeholder and the strong supervision signal.

	\begin{table}
		\caption{\qiankun{The impact of the ratio between the number of negative and positive samples on re-identification capability. These models are trained on CrowdHuman \cite{shao2018crowdhuman} and evaluated on MOT17 train split without fine-tuning. }}
		\setlength{\tabcolsep}{10pt}
		\centering
		\scriptsize
		\begin{tabular}{c|c|c|c}
			\hline
			Tracker & $N^{neg}/N^{pos}$  &R1$\uparrow$ &mAP $\uparrow$ \\
			\hline
			\multirow{6}*{FairMOT+UTrack}
			&1/9 &54.6\% &32.8\% \\
			&2/8 &56.4\% &\underline{\bf34.1\%} \\
			&3/7 &\underline{\bf58.1\%} &31.8\% \\
			&4/6 &57.9\% &31.3\% \\
			\cline{3-4}
			&5/5 &\multicolumn{2}{c}{not converged} \\
			& 6/4 &\multicolumn{2}{c}{not converged} \\
			\hline
		\end{tabular}
		\label{table_pretrain_on_crowdhuman_no_ft_reid_neg_pos}
	\end{table}

	\qiankun{We then evaluate the impact of $\frac{N^{neg}}{N^{pos}}$ on re-identification capability while training on image-based data in \tref{table_pretrain_on_crowdhuman_no_ft_reid_neg_pos}. The model achieves the best R1 score when $\frac{N^{neg}}{N^{pos}}=\frac{3}{7}$, but achieves the best mAP score when $\frac{N^{neg}}{N^{pos}}=\frac{2}{8}$. Interestingly, we find that the Re-ID module cannot converge if $\frac{N^{neg}}{N^{pos}} \ge 1$. We set  $\frac{N^{neg}}{N^{pos}}$ to $\frac{2}{8}$ for the balance between R1 and mAP scores.}

	In \tref{table_pretrain_on_crowdhuman_no_ft}, we show the tracking results  of trackers on MOT17 dataset by directly applying the CrowdHuman pre-trained model without fine-tuning. Compared with the supervised Re-ID learning in FairMOT and unsupervised Re-ID learning CysAs \cite{wang2020cycas}, our UTrack performs much better in terms of MOTA, IDF1, MT, FN and IDS, demonstrating the superiority of the proposed unsupervised Re-ID learning method. 
	
	\begin{table}
		\caption{Tracking results on the MOT17 validation set by directly applying the pre-trained trackers on CrowdHuman \cite{shao2018crowdhuman} without fine-tuning. 
		}
		\setlength{\tabcolsep}{1.7pt}
		\centering
		\scriptsize
		\begin{tabular}{c|c|c|c|c|c|c|c}
			\hline
			Trackers  &MOTA$\uparrow$ &IDF1 $\uparrow$ &MT$\uparrow$ &ML$\downarrow$ &FP$\downarrow$ &FN $\downarrow$ &IDS$\downarrow$ \\
			\hline
			FairMOT \cite{zhang2020fairmot} &64.0\% &64.6\% &138 &63 &\underline{\bf2130} &16806 &501 \\
			FairMOT+CysAs \cite{wang2020cycas} &63.9\% &64.9\% &137 &\underline{\bf62} &2781 &16105 &594  \\
			FairMOT+UTrack &\underline{\bf64.8\%} &\underline{\bf69.2\%} &\underline{\bf143} &64 &2390 &\underline{\bf16203} &\underline{\bf418}  \\
			\hline
		\end{tabular}
		\label{table_pretrain_on_crowdhuman_no_ft}
	\end{table}

	\begin{table}
		\caption{Tracking results on the MOT17 validation set by fine-tuning the CrowdHuman pre-trained trackers on the MOT17 dataset. Trackers marked with/without $\star$ correspond pre-training on CrowdHuman or not.
		}
		\setlength{\tabcolsep}{1.2pt}
		\centering
		\scriptsize
		\begin{tabular}{c|c|c|c|c|c|c|c}
			\hline
			trackers  &MOTA$\uparrow$ &IDF1 $\uparrow$ &MT$\uparrow$ &ML$\downarrow$ &FP$\downarrow$ &FN $\downarrow$ &IDS$\downarrow$  \\
			\hline
			\multicolumn{8}{c}{impact on Re-ID module}\\
			\hline
			FairMOT \cite{zhang2020fairmot} &67.5\% &70.2\% &134 &55 &2814 &14263 &492 \\
			FairMOT$^\star$ \cite{zhang2020fairmot} &70.7\% &\underline{\bf74.7\%} &\underline{\bf172} &48 &3255 &12171 &\underline{\bf431} \\
			\hline
			FairMOT+CysAs \cite{wang2020cycas} &67.0\% &70.8\% &134 &60 &2631 &14681 &503  \\
			FairMOT+CysAs \cite{wang2020cycas}$^\star$ &69.4\% &70.8\% &158 &46 &3412 &12515 &592  \\
			\hline
			FairMOT+UTrack &67.6\% &71.8\% &137 &63 &\underline{\bf2621} &14388 &503 \\
			FairMOT+UTrack$^\star$  &\underline{\bf70.8\%} &73.8\% &165 &\underline{\bf45} &3222 &\underline{\bf12052} &524 \\
			\hline
			\hline
			\multicolumn{8}{c}{impact on occlusion estimation module}\\
			\hline
			FairMOT+UTrack+OccE &68.5\% &72.0\% &142 &57 &2840 &13797 &\underline{\bf396} \\
			FairMOT+UTrack+OccE$^\star$ &\underline{\bf72.0\%} &\underline{\bf73.1\%} &\underline{\bf168} &\underline{\bf46} &3565 &\underline{\bf11119} &417 \\
			\hline
			CenterTrack\cite{zhou2020tracking}  &60.7\% &62.7\% &112 &76 &\underline{\bf2179} &18447 &564 \\
			CenterTrack$^\star$\cite{zhou2020tracking}  &66.1\% &64.2\% &140 &72 &2442 &15286 &588 \\
			\hline
			CenterTrack+OccE  &62.1\% &64.6\% &127 &68 &3372 &16583 &440 \\
			CenterTrack+OccE$^\star$  &67.4\% &67.6\% &158 &63 &4086 &13107 &414 \\
			\hline
		\end{tabular}
		\label{pretrain}
	\end{table}	
	
	We further show the results in \tref{pretrain} by fine-tuning the CrowdHuman pre-trained models (marked by $\star$) on the MOT17 dataset. 
	For reference, we also provide the results without pre-training. From the results, we can observe that pre-training on the image-based data can generally boost the overall tracking performance. Compared to the supervised Re-ID learning method used in FairMOT, our unsupervised tracker UTrack can achieve very comparable tracking performance without using any ID supervision. By contrast, pre-training on image-based data using CysAs cannot improve the IDF1 performance.
	
	\begin{table*}[tp]
		\caption{Benchmark results on MOTChallenge. Trackers marked with $\dagger$ track objects in an offline manner.  FairMOT$^\S$ is pre-trained on CrowdHuman without five extra datasets\textsuperscript{\ref{five_extra_datasets}}. Best results are shown in \textbf{bold} and highlighted with underline.
		}
		\setlength{\tabcolsep}{0.5pt}
		\centering
		\scriptsize
		\begin{tabular}{c|c|c|c|c|c|c|c|c|c|c|c|c}
			\hline
			\multicolumn{2}{c|}{benchmark}  &methods  
			& MOTA$\uparrow$ & IDF1$\uparrow$ & MT$\uparrow$ & ML$\downarrow$ & FP$\downarrow$ & FN$\downarrow$ &Recall$\uparrow$ & IDS$\downarrow$ & Frag$\downarrow$  & Hz$\uparrow$ \\ 
			\hline
			\multirow{13}*{\rotatebox{90}{MOT16}}
			& \multirow{6}*{\rotatebox{0}{Public}} 
			& Tracktor++ \cite{bergmann2019tracking}& 56.2\% & 54.9\% & 20.7\% & 35.8\% & \underline{\bf2394} & 76844 &57.9\% & 617 & 1068 & 1.6 \\ 
			&   & GSM$_{\rm Tracktor}$ \cite{liugsm} & 57.0\% & 58.2\%  & 22.0\% & 34.5\% & 4332 & 73573 &59.6\% & 475 & 859  & 7.6 \\ 
			&   & MPNTrack$^\dagger$ \cite{braso2020learning} & 58.6\% & 61.7\% & 27.3\% & 34.0\% & 4949 & 70252 &61.5\% & \underline{\bf354} & \underline{\bf684}  & 6.5 \\
			&   & Lif\_T$^\dagger$ \cite{hornakova2020lifted} & 61.3\% & 64.7\%  & 27.0\% & 34.0\% & 4844 & 65401 &64.1\% & 389 & 1034 & 0.5 \\
			& & TMOH \cite{stadler2021improving} &63.2\% &63.5\% &27.0\% &31.0\% &3122 &63376 & 65.2\% &635 &1486 & 0.7 \\
			&   &OTrack$_{\rm ct}$ (ours) & 65.3\% & 62.7\%  &26.1\% & 34.9\% &5179 &57484 &68.5\%  &628 &1616  & 17.2 \\  
			&   & OUTrack$_{\rm fm}$ (ours) & \underline{\bf69.3\%} & \underline{\bf67.5\%}  &\underline{\bf37.3\%} &\underline{\bf19.1\%} &10657 & \underline{\bf44059} &\underline{\bf75.8\%}  &1284 &2677  & \underline{\bf24.5}  \\ 
			\cline{2-13}
			& \multirow{7}*{\rotatebox{0}{Private}}
			& JDE\cite{wang2019towards} & 64.4\% & 55.8\%  & 35.4\% & 20.0\% & - & - &- & 1544 & - & 22.0 \\ 
			&   & LM\_CNN$^\dagger$ \cite{babaee2019dual} &67.4\% & 61.2\% & 38.2\% & 19.2\% & 10109 & 48435 &73.4\% & 931 & 1034 & 1.7 \\ 
			&   & LMP$^\dagger$ \cite{tang2017multiple} & 71.0\% & \underline{\bf80.2\%} & \underline{\bf46.9\%} & 21.9\% & \underline{\bf7880} & 44564 &75.6\% & \underline{\bf434} & \underline{\bf587} & 0.5 \\ 
			& &SOTMOT \cite{zheng2021improving} & 72.1\% &72.3\% &44.0\% &13.2\% &14344 &34784 &- &1681 &- &16 \\
			&   & FairMOT\cite{zhang2020fairmot} & \underline{\bf74.9\%} &72.8\% & 44.7\% & 15.9\% & 10163 & 34484 &81.1\% & 1074 & 2567 & 25.4 \\
			&   & FairMOT$^\S$\cite{zhang2020fairmot} & 72.7\% &74.0\% & 42.0\% & 17.8\% & 12930 &35804 &80.4\% & 1121 & 2732 & \underline{\bf25.4} \\
			&   & OTrack$_{\rm ct}$ (ours) &73.3\% &70.3\%  &41.3\% &15.9\% &30057 & 115944 &79.5\% &4440 &8742 & 17.0  \\ 
			&   & OUTrack$_{\rm fm}$ (ours) &74.2\% & 71.1\%  & 44.8\% & \underline{\bf14.0\%} & 13214 & \underline{\bf32581} &\underline{\bf82.1\%} &1324 &2413 & 24.8  \\ 
			\hline
			\multirow{13}*{\rotatebox{90}{MOT17}} 
			& \multirow{8}*{\rotatebox{0}{Public}}
			& Tracktor++ \cite{bergmann2019tracking} & 56.3\% & 55.1\%  & 21.1\% & 35.3\% & \underline{\bf8866} & 235449 &58.3\% & 1987 & 3763 & 1.5 \\ 
			&   & GSM$_{\rm Tracktor}$ \cite{liugsm} & 56.4\% & 57.8\%  &22.2\% & 34.5\% & 14379 & 230174 &59.2\% & 1485 & 2763 & 8.7 \\ 
			&   & MPNTrack$^\dagger$ \cite{braso2020learning} & 58.8\% & 61.7\% & 28.8\% & 33.5\% & 17413 & 213594 &62.1\% & \underline{\bf1185} &\underline{\bf2265} & 6.5 \\ 
			&   & Lif\_T$^\dagger$ \cite{hornakova2020lifted} & 60.5\% & 65.6\% & 27.0\% & 33.6\% & 14966 & 206619 &63.4\%  & 1189 & 3476 & 0.5 \\ 
			&   & CenterTrack \cite{zhou2020tracking} & 61.5\% & 59.6\% & 26.4\% &31.9\% &14076 & 200672 &64.4\% & 2583 & 4965 & 17.5 \\ 
			& & TMOH \cite{stadler2021improving} &62.1\% &62.8\% &26.9\% &31.4\% &10951 &201195 & 64.3\% &1897 &4622 & 0.7 \\
			& &SiamMTOT \cite{shuai2021siammot} & 65.9\% & 63.3\% &34.6\% &23.9\%  	&18098 & 170955 & - & - &- &17 \\		
			
			&   & OTrack$_{\rm ct}$ (ours) &63.9\% &62.3\% &25.7\% &35.5\% &14903 &186878 &66.9\% &1949 & 4952  &17.2  \\
			&   & OUTrack$_{\rm fm}$ (ours) & \underline{\bf69.0\%} &\underline{\bf66.8\%} &\underline{\bf37.6\%} &\underline{\bf19.7\%} &28855 &\underline{\bf141587} &\underline{\bf74.9\%} &4449 &8733  &\underline{\bf24.8} \\
			\cline{2-13}
			& \multirow{5}*{\rotatebox{0}{Private}}
			& CenterTrack \cite{zhou2020tracking} & 67.8\% & 64.7\% & 34.6\% &24.6\% & \underline{\bf18498} & 160332 &71.6\% & 3039 & 6102  & 17.5 \\ 
			& &SOTMOT \cite{zheng2021improving} & 71.0\% &71.9\% &42.7\% &15.3\% &39537 &118983 &- &5184 &- &16 \\
			&   & FairMOT \cite{zhang2020fairmot} & \underline{\bf73.7\%} &72.3\% & 43.2\% & 17.3\% & 27507 & 117477 &79.2\% & 3303  & 8073 & \underline{\bf25.9} \\
			&   & FairMOT$^\S$ \cite{zhang2020fairmot} & 71.8\% & \underline{\bf73.1\%} & 40.9\% & 19.0\% & 34764 & 120909 &78.6\% & 3534  & 8724 & \underline{\bf25.9} \\
			&   & OTrack$_{\rm ct}$ (ours) &69.0\% &67.8 \%  & 35.4\% &21.1\% &39159 &133143 &76.4\% &\underline{\bf2643} & 6261 & 16.9 \\ 
			&   & OUTrack$_{\rm fm}$ (ours) &73.5\% & 70.2\%  & \underline{\bf43.3\%} & \underline{\bf15.2\%} & 34764 & \underline{\bf110577} &\underline{\bf80.4\%} & 4110 & 7506 & 25.4 \\ 
			\hline
			\multirow{9}*{\rotatebox{90}{MOT20}} 
			& \multirow{5}*{\rotatebox{0}{public}}
			& SORT \cite{bewley2016simple} & 42.7\% & 45.1\% & 16.7\% & 26.2\% & 27521 & 264694 &48.8\% & 4470 & 17798 & 57.3 \\ 
			& & MLT\cite{zhang2020multiplex} & 48.9\% & 54.6\% & 30.9\% & 22.1\% & 45660 & 216803 &58.1\% & 2187 & 3067 & 3.7 \\ 
			& &Tracktor++\cite{bergmann2019tracking} & 52.6\% & 52.7\% & 30.3\% & 25.0\% & \underline{\bf6439} & 36680 &55.4\% & \underline{\bf1648} & 4374 & 1.2 \\
			& & TMOH \cite{stadler2021improving} &60.1\% &61.2\% &46.7\% &17.8\% &38043 &165899 & 67.9\% &2342 &4320 & 0.6 \\
			&   &OTrack$_{\rm ct}$ (ours) & 60.8\% &60.0\% &\underline{\bf56.4\%} &14.3\% &70156 &129703 &\underline{\bf74.9\%} &2783 &\underline{\bf2984} & 9.4 \\ 
			&   & OUTrack$_{\rm fm}$ (ours) &\underline{\bf65.3\%} &\underline{\bf65.0\%} &49.4\% &\underline{\bf13.3\%} &38709 &\underline{\bf13799} &73.4\% &2832 & 7212 & \underline{\bf10.7} \\ 
			\cline{2-13}
			& \multirow{4}*{\rotatebox{0}{private}}
			& FairMOT \cite{zhang2020fairmot} & 61.8\% & 67.3\%  & \underline{\bf68.8\%} &\underline{\bf7.6\%} & 103440 & \underline{\bf88901} &\underline{\bf80.6\%} & 5243 & 7874  &\underline{\bf13.2} \\ 
			& &FairMOT$^\S$ \cite{zhang2020fairmot} & 68.1\% & 71.1\%  & 53.3\% &12.9\% &  \underline{\bf30503} & 131380 &74.6\% & 3019 & 10509  &\underline{\bf13.2} \\ 
			& &SOTMOT \cite{zheng2021improving} & \underline{\bf68.6\%} &\underline{\bf71.4\%} &64.9\% &9.7\% &57064 &101154 &- &4209 &- &8.5 \\
			&   & OTrack$_{\rm ct}$ (ours) & 65.8\% & 61.8\%  &58.9\% &13.1\% &48947 & 125152 &73.8\% & 2897 &\underline{\bf3064} & 10.3 \\ 
			&   & OUTrack$_{\rm fm}$ (ours) & 68.5\% &69.4\%  &57.9 \% & 12.2\% & 37431 & 123197 &76.2\% & \underline{\bf2147} & 5683 & 12.4 \\ 
			\hline
		\end{tabular}
		\label{table_mot}
	\end{table*}
	
	\subsection{Results on MOTChallenge}
	\label{section_results_on_motchallenge}
	Though the main focus of this paper is not to achieve the state-of-the-art performance, we still show the tracking results on the standard MOTchallenge benchmark by integrating the proposed modules into FairMOT \cite{zhang2020fairmot} and CenterTrack \cite{zhou2020tracking} respectively.
	For notation simplicity, we denote the variant of FairMOT that integrates our unsupervised Re-ID learning and occlusion estimation module as "OUTrack$_{\rm fm}$", and the variant of CenterTrack that integrates our occlusion estimation module as "OTrack$_{\rm ct}$". We conduct the evaluation on both public and private detection. 
	For the public detection evaluation, we follow the works in \cite{bergmann2019tracking, liugsm, zhou2020tracking, he2021learnable, kim2021discriminative, saleh2021probabilistic, wang2021multiple, wu2021track} to refine the public detections and keep the bounding boxes that are close to the tracked objects.
	Note that only the provided training sequences are used to train the model for public detection evaluation. For private detection, we follow CenterTrack and FairMOT to pre-train our tracker with the CrowdHuman dataset. However, the original FairMOT also involves a mixture dataset that consists of five extra datasets\footnote{These datasets are ETH \cite{ess2008mobile}, CityPerson \cite{zhang2017citypersons}, CalTech \cite{dollar2009pedestrian}, CUHK-SYSU \cite{xiao2017joint} and PRW \cite{zheng2017person}. Besides the box annotations, the identity information is also provided in the last three datasets. \label{five_extra_datasets} }. For fair comparison, we train FairMOT on CrowdHuman and MOTChallenge without these five extra datasets, which is denoted as FairMOT$^\S$.
	
	Results are shown in \tref{table_mot}. Overall, our FairMOT based tracker OUTrack$_{\rm fm}$ achieves state-of-the-art performance on all datasets and our CenterTrack based tracker OTrack$_{\rm ct}$ outperfoms CenterTrack by a large margin. Compared to offline methods, such as Lif\_T \cite{hornakova2020lifted} and MPNTrack \cite{braso2020learning}, both OUTrack$_{\rm fm}$ and OTrack$_{\rm ct}$  have a higher Frag, the reason is that the short broken tracklets can be linked to a long trajectory by a post process, which is not allowed in online trackers.
	
	Compared with FairMOT, OUTrack$_{\rm fm}$ achieves very comparable performance on MOT17 with much less pretraining data and better performance with the same pretraining data in terms of MOTA. As for IDF1, FairMOT performs better  than OUTrack$_{\rm fm}$ by 2.1\%. The main reason is that FairMOT supervises Re-ID module with identity information, while the Re-ID module in OUTrack$_{\rm fm}$ is trained in a totally unsupervised manner. 
	It is interesting that OUTrack$_{\rm fm}$ performs much better than FairMOT on MOT20 with private detection. The main reasons may be in two folds: 1) FairMOT fine-tunes the models on MOT20 after pre-training on the mixture dataset, while the mixed datasets used in FairMOT are different from MOT20 that is captured in crowd scenes. However, we fine-tune the model on MOT20 after pre-training on CrowdHuman and the images in CrowdHuman are all collected from crowd scenes. 2) A higher detection confidence ($0.5$) is used in OUTrack$_{\rm fm}$, since FairMOT ($0.3$) has a much higher FP. When training FairMOT on the same dataset as OUTrack$_{\rm fm}$ and increasing the detection confidence to $0.5$, the main metric MOTA is greatly improved. But OUTrack$_{\rm fm}$ still performs better.

	\begin{figure*}[t]
		\centering
		\includegraphics[width=1.0\columnwidth]{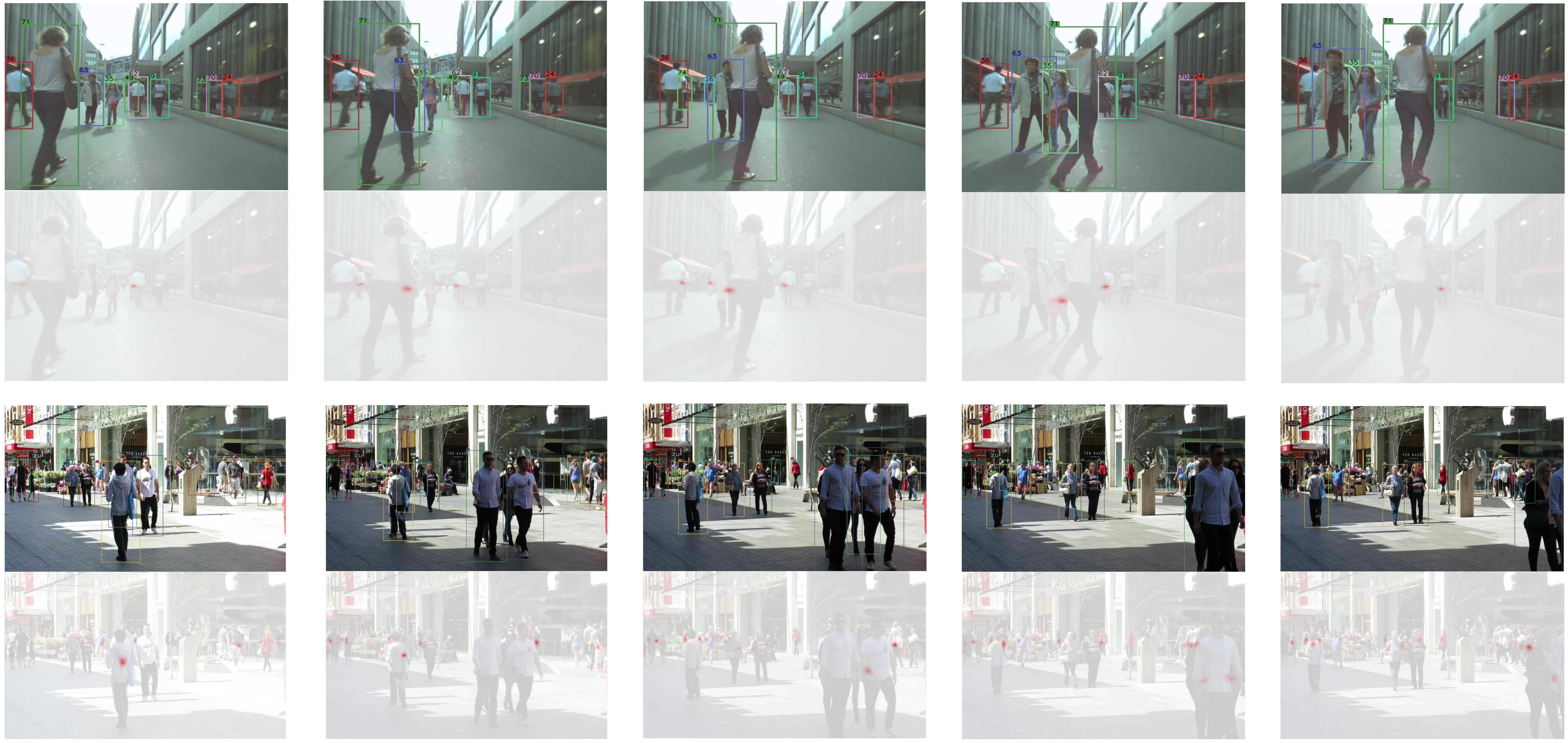}
		\caption{
			The tracking results and predicted occlusion heatmaps of OUTrack$_{\rm fm}$.  
		}
		\label{figure_tracking_results}
	\end{figure*}

	Compared to CenterTrack, OTrack$_{\rm ct}$ performs much better on MOT17 for both the private and public detection settings. For example, our OTrack$_{\rm ct}$ surpasses CenterTrack on MOT17 in terms of MOTA by $2.4\%$ and $1.2\%$ with public and private detection, respectively. 
	Through finding the lost objects based on the predicted occlusions, a higher Recall is also achieved. Though more FP are involved when finding the missed objects, the FN is greatly reduced.

	In \fref{figure_tracking_results}, some tracking results and the predicted occlusion heatmaps of OUTrack$_{\rm fm}$ are presented. For the first case, we can see that some objects (63 and 35) can be re-identified when they reappeared after a short-term disappearance, demonstrating the effectiveness of our unsupervised Re-ID learning method. For both cases, occlusions between different objects can be effectively detected by occlusion estimation module and those highly occluded objects still can be tracked, indicating the effectiveness of our occlusion estimation module.
	
	\section{Conclusion}
	\label{section_conclusion}
	In this paper, we present a new occlusion-aware multi-object tracking framework. It involves two key modules: unsupervised Re-ID learning and occlusion estimation module. The unsupervised Re-ID learning adopts an unsupervised matching based loss between adjacent frames, whose motivation is that objects with the same identity in adjacent frames share similar appearance and objects in
	two images that from different scenes (or within the same image) have
	different identities and appearances. Compared to the supervised classification based Re-ID learning, it does not suffer from the dimension explosion issue for a large identity number and is more friendly to real large-scale applications. The occlusion estimation module can alleviate the tracking lost issue caused by missing detection. It can find the occluded objects back by estimating the occlusion map that shows all possible occlusion locations.
	The two proposed modules can be applied to existing MOT systems in a natural way and demonstrate their effectiveness. 
	
	\section{Acknowledgement}
	This  work  is  supported  by  the  National  Natural  Science Foundation  of China (No. U20B2047, No. 62002336).

	\bibliographystyle{elsarticle-harv} 
	\bibliography{cite}

\end{document}